\documentclass[letterpaper, 10 pt, conference]{ieeeconf}  

\IEEEoverridecommandlockouts                              
\usepackage{mathtools}
\usepackage{comment}
\usepackage{graphicx}
\usepackage{amsmath}
\usepackage{amssymb}
\usepackage{svg}
\usepackage{pdfpages}
\usepackage{caption}
\usepackage{subcaption}
\usepackage{placeins}
\usepackage{cleveref}

\title{\LARGE \bf
Hyper-GST: Predict Metro Passenger Flow Incorporating GraphSAGE, Hypergraph, Social-meaningful Edge Weights and Temporal Exploitation }

\author{
  Yuyang, Miao\\
  \texttt{yuyang.miao20@imperial.ac.uk}
  \and
  Yao, Xu\\
  \texttt{yao.xu15@imperial.ac.uk}
  \and
  Danilo, Mandic\\
  \texttt{d.mandic@imperial.ac.uk}
}

\begin{document}

\maketitle
\thispagestyle{empty}
\pagestyle{empty}

\begin{abstract}

Predicting metro passenger flow precisely is of great importance for dynamic traffic planning. Deep learning algorithms have been widely applied due to their robust performance in modelling non-linear systems. However, traditional deep learning algorithms completely discard the inherent graph structure within the metro system. Graph-based deep learning algorithms could utilise the graph structure but raise a few challenges, such as how to determine the weights of the edges and the shallow receptive field caused by the over-smoothing issue. To further improve these challenges, this study proposes a model based on GraphSAGE with an edge weights learner applied. The edge weights learner utilises socially meaningful features to generate edge weights. Hypergraph and temporal exploitation modules are also constructed as add-ons for better performance. A comparison study is conducted on the proposed algorithm and other state-of-art graph neural networks, where the proposed algorithm could improve the performance.

\end{abstract}

\section{INTRODUCTION}

Traffic flow forecasting is a necessity for modern intelligent traffic systems. The precise prediction of traffic flow could help route planning, signal coordination and reducing traffic hazards such as congestion. Among all the traffic flow prediction tasks, metro ridership prediction has drawn tremendous interest in academic and industry domains due to its vital position in the city transportation network. 

Statistical models have played an important role in traffic flow prediction since the last century. Levin and Tsao concluded that ARIMA (0,1,1) model work best on traffic flow prediction task \cite{levin1980forecasting}. Moayedi and Masnadi-Shirazi used ARIMA to predict passenger flow and detect anomalies from normal traffic variation \cite{moayedi2008arima}. Variations of ARIMA have also been applied to traffic prediction tasks. Otoshi et al. combined ARIMA and seasonal ARIMA to model both short-term and long-term traffic flow to give a better prediction. \cite{otoshi2015traffic}. Zhou et al. combined the linear model ARIMA and the non-linear model GARCH to achieve prediction in both long and short scales.

However, the typical statistical model completely excludes the dependency between different time series and the information contained. On the other hand, deep learning is powerful in modelling non-linear processes and can extract high-level features for the whole traffic flow dataset. Deep learning has become a promising tool for metro ridership prediction with the explosion of computation power and advances in data collection and storage techniques. The Convolutional Neural Network (CNN) and Long-short Time memory (LSTM) modules are used separately or combined to capture the spatial and temporal information \cite{wang2018detection,yu2016data,shao2021traffic,narmadha2021spatio,ranjan2020city}. 

One major flaw of the traditional deep learning methods relies upon 2D Euclidean space and works on data in the regular domain. On the contrary, Graph Convolutional Network (GCN) and its extension proposed recently could utilise the inherent graph structure in metro transportation \cite{kipf2016semi} \cite{hamilton2017inductive} \cite{velivckovic2017graph}. Ali, Zhu and Zakarya combined GCN and LSTM to build a spatial-temporal graph neural network \cite{ali2022exploiting}. Agafonov applied convolution both in the graph and temporal domain to predict the traffic flow\cite{agafonov2020traffic}. However, there are still some issues that need to be solved.
\begin{itemize}
    \item The edge weights scheme of the metro graph. The ridership of each line is closely related to social factors such as population and housing prices. Thus, an unweighted graph or a graph using distances as edge weights could not fully express the natural physical process.
    \item The structure of the metro graph. The original metro graph has a restrictive receptive field. The receptive field can only be increased by stacking layers which might cause over-smoothing. Thus, it is necessary to redefine the structure of the metro graph.
    \item Most work focus on mimicking the passenger flow between stations. However, the passengers' behaviour of changing metro lines is also worth investigating.
\end{itemize}

To improve on these challenges above, we proposed a GraphSAGE-based model. The main body of the algorithm contains a GraphSAGE convolution module with an edge weights learner. The edge weights learner takes the social features of two stations and produces the corresponding edge weights. The original metro graph is transformed into a $k$-hop graph. Then the $k$-hop graph is fed into the GraphSAGE module, and the edge weights learner will assign weights to the $k$-hop graph. A hypergraph and a temporal exploitation module will also be constructed along with the GraphSAGE main body. The hypergraph is designed to mimic the passengers' changing metro line behaviour, and the temporal exploitation module is to extract information in the temporal domain. The hypergraph module and temporal exploitation module will be considered to be add-ons. When selected, the add-on output will be concatenated with the output of the main body and produce the final prediction through fully connected layers. In summary, the proposed module could contribute to the following:

\begin{itemize}
    \item A model could selectively fuse information from temporal and hypergraph domains.
    \item A new way to learn edge weights based on social-meaningful features
    \item A model that could extract metro line changing information 
    \item A way of extracting information from hypergraph.
\end{itemize}

The rest of the paper is organised as follows: In section II, we will review the related works, namely the edge weights schemes and the use of deep learning for traffic flow prediction. In section III, we will give the methodology of our proposed study. Firstly, we introduce the dataset and then explain its idea and mechanism in detail. Section IV represents the results of the proposed method and compares it with the state-of-art models. Finally, Section V discuss the results.   

\section{RELATED WORK}

\subsection{Deep learning for traffic flow prediction}
With the rapid growth of computation power and data, deep learning has been widely applied to traffic flow prediction. LSTM is a popular choice due to their ability to model temporal dependencies. Yang et al. improved the performance of LSTM by learning enhanced traffic feature, which is formed by combining features at different time scales \cite{yang2019traffic}. Cai et al. proposed a noise-immune LSTM model for traffic flow prediction by employing the correntropy criterion as the loss function \cite{cai2020noise}. Fang et al. used a Kalman filter to adjust the output of LSTM further, achieving real-time update of state variable \cite{fang2021kalman}. 

While LSTM-based methods could achieve good performance, they completely ignore the spatial information contained in a traffic network. Some researchers utilised CNN to extract spatial information. Yu et al. applied a 3-D Convolutional Network for large-scale traffic prediction \cite{yu2016data}. Mehdi et al. used CNN on meta-parameters to predict the congestion state, which is determined based on entropy labelling \cite{mehdi2022entropy}. Nguyen et al. applied equilibrium optimiser instead of backpropagation to train a CNN to predict traffic transportation \cite{nguyen2020eo}. 

There are also works combining LSTM and CNN to model both temporal and spatial dependencies. Narmadha and Vijayakumar firstly used a stacked autoencoder to denoise the multivariate traffic data. Then the denoised data was fed through a CCN and LSTM to extract spatial and temporal features \cite{narmadha2021spatio}. Li et al. firstly decomposed traffic data using wavelet decomposition. Then CNN-LSTM block was applied to different levels of decomposed data. Finally, the prediction was recovered from the wavelet results \cite{li2021hybrid}. 

\subsection{Edge Weights}
Unlike grid data in the regular domain, the graph's structure contains information that might be useful. Thus, a proper edge weights scheme can help improve the model's performance. Some researchers defined their edge weight scheme by directly using or learning from geometric and traffic data. Chen et al. used the transition values of travel times as edge weights \cite{chen2018dynamic}. Li et al. defined the edge weight between two stations as the number of travels between them \cite{li2018graph}. Huang et al. directly regarded distances between stations as edge weights \cite{huang2020lsgcn}. However, in the context of metro transport, there is no direct connection between the importance of the route and the route length or the travel time.

It is also a popular choice to learn weight from the data itself. Zhe et al. applied fast time warping to calculate the distance between the in or outflow at each segment and took the similarity as edge weights \cite{chen2020multitask}. Chen et al. transformed the graph signal into spectral space and learned the edge weights by feeding the coordinates difference of two nodes in spectral space to a kernel \cite{chen2021graph}. Lu et al. calculated dynamic attention scores from nodes to retrieve dynamic edge weights \cite{lu2020spatiotemporal}. Yu et al. defined an edge weight scheme for each layer of the GCN network. There are parameters to be learned for each layer, and each layer's edge weights are calculated using that layer's parameters, the distance between two stations and the number of lanes at each node \cite{yu2020forecasting}. Chen et al. calculated edge weights using a Gaussian kernel based on two node's degrees  \cite{chen2020multi}. He et al. compared the results of GAT and GATv2n and showed that GATv2n could improve GAT's results. Both models calculated edge weights based on the attention scores calculated from the graph data itself. These models could improve performance by assigning attention scores as weights. However, there is no real-world meaning of these weights.

For this study, traditional deep learning algorithms will be discarded, and graph-based deep learning algorithms will be employed to utilise the inherent graph structure of the metro system. Also, unlike other works, we will build the edge weights based on social meaningful features. The following methodology chapter will give a detailed explanation.

\section{Methodology}
This section, we will introduce the dataset, and elaborate the details of our proposed model.

\subsection{Dataset}
The dataset is from \emph{VIS-tube} project \cite{data}. The dataset contains traffic data, geometrical data and social data. The traffic data contains \textbf{Entry} and \textbf{Exit} data for each station at five time stamps, namely \textbf{Early}, \textbf{AM}, \textbf{Mid}, \textbf{PM} and \textbf{Late}. The traffic data is based on yearly average scale and it includes 13 years, \textbf{2003-2012} and \textbf{2014-2016}. For this study, stations within Zone 4 (206 in total) are used. Thus the traffic data has a overall shape of $\mathbf{206 \times 10 \times 13}$.
\begin{figure}[h]
    \centering
    \includegraphics[width = 1 \linewidth]{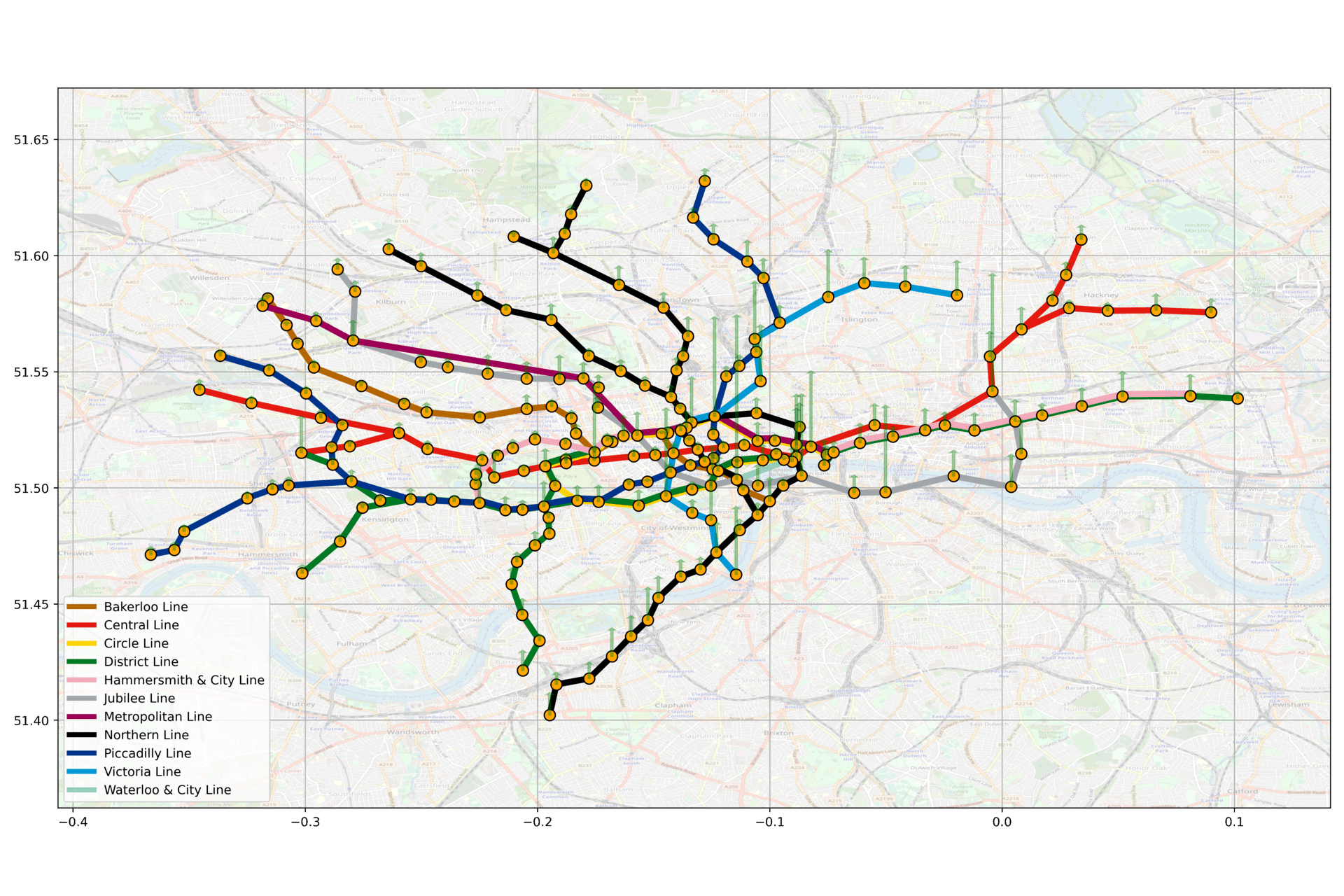}
    \caption{The AM entry data for the year 2016.}
    \label{fig:2016_am_entry}
\end{figure}

The dataset also has social features. For this study, the zone of, average housing price and expected life span around each station is used for edge weights learning.

\begin{figure}[h]
    \centering
    \includegraphics[width = 1 \linewidth]{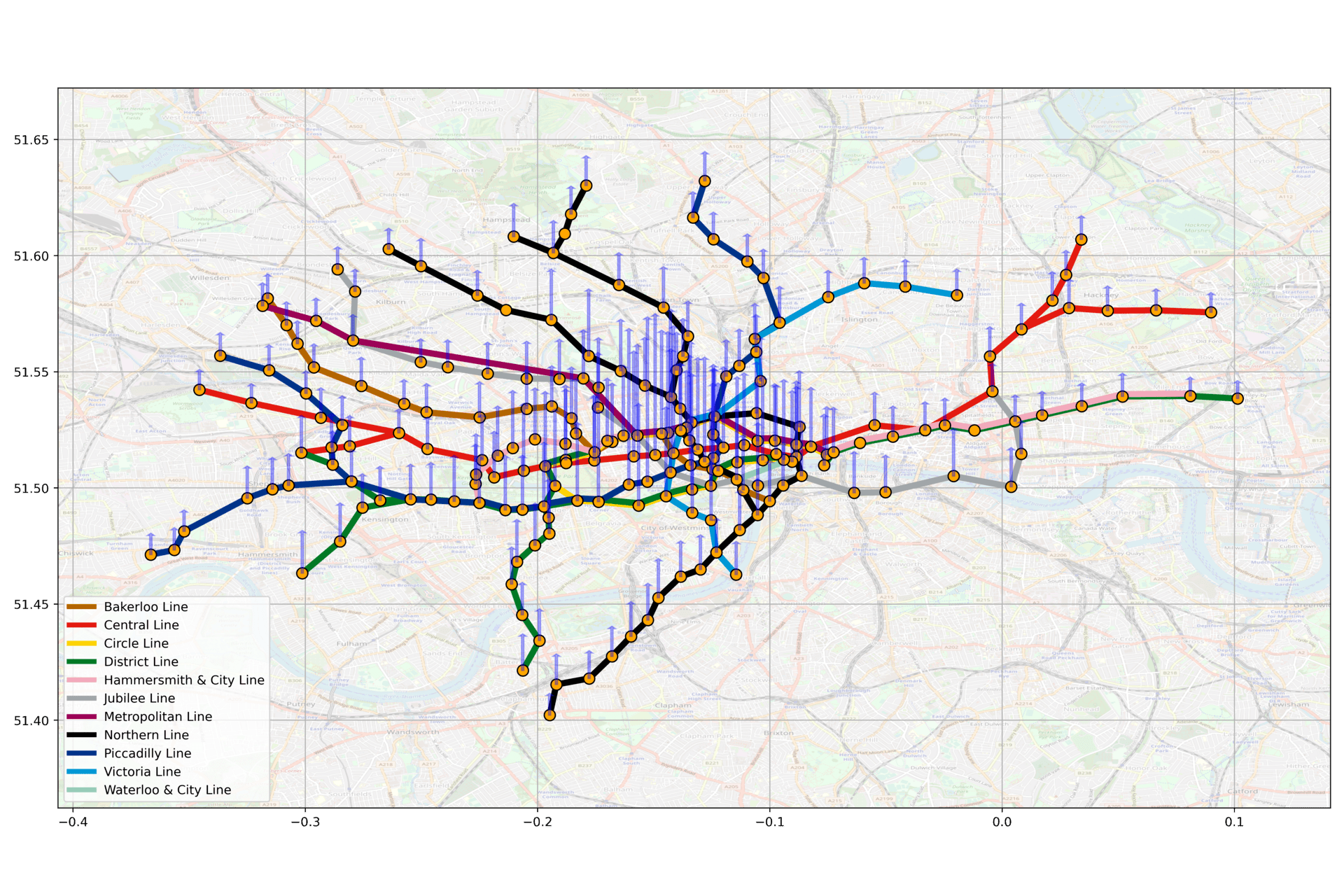}
    \caption{The average housing price around each station.}
    \label{fig:2016_am_entry}
\end{figure}

\begin{figure}[h]
    \centering
    \includegraphics[width = 1 \linewidth]{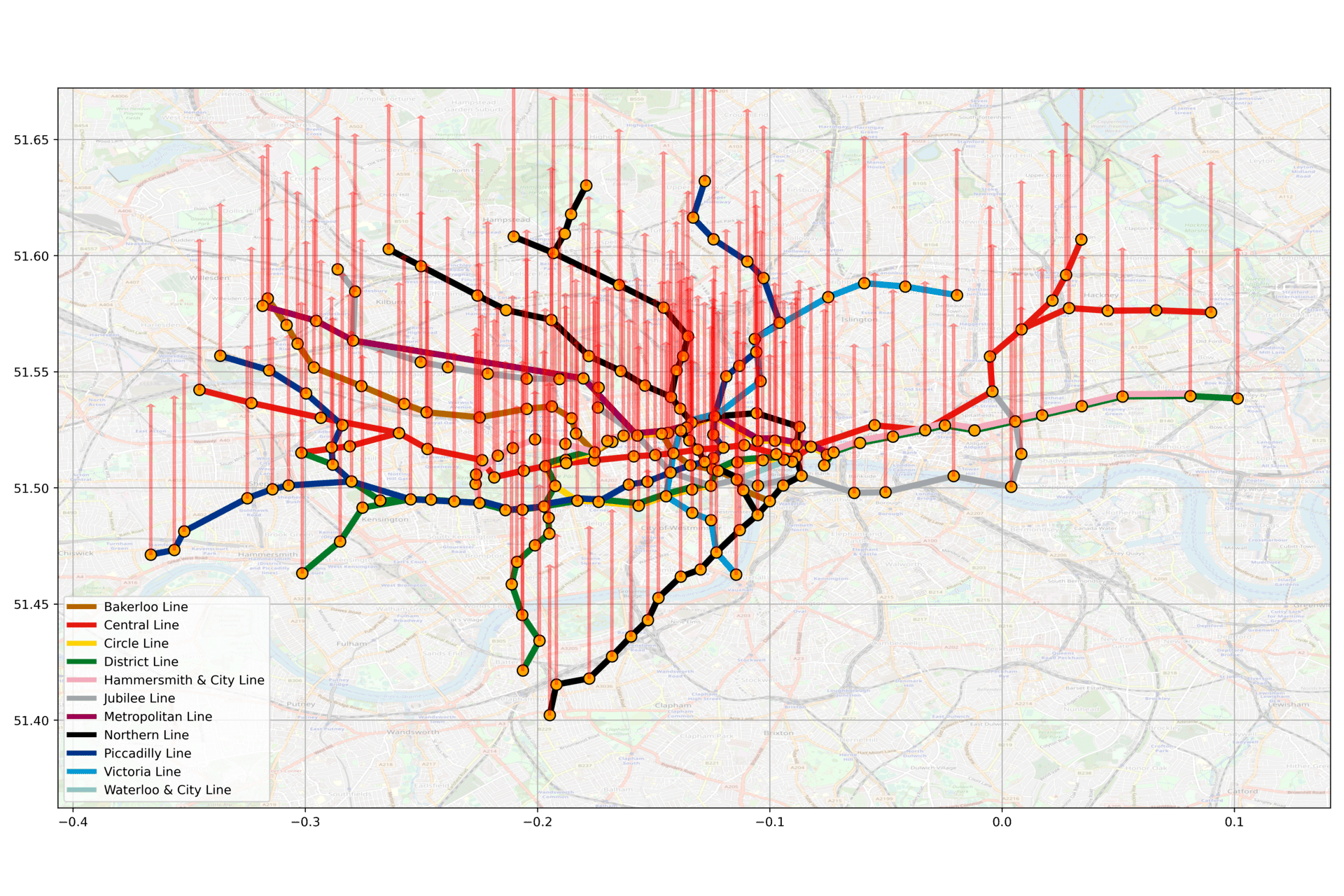}
    \caption{The average expected life span around each station.}
    \label{fig:2016_am_entry}
\end{figure}

\subsection{Problem Formulation}
\begin{figure}[h]
  \includegraphics[width = 1 \linewidth]{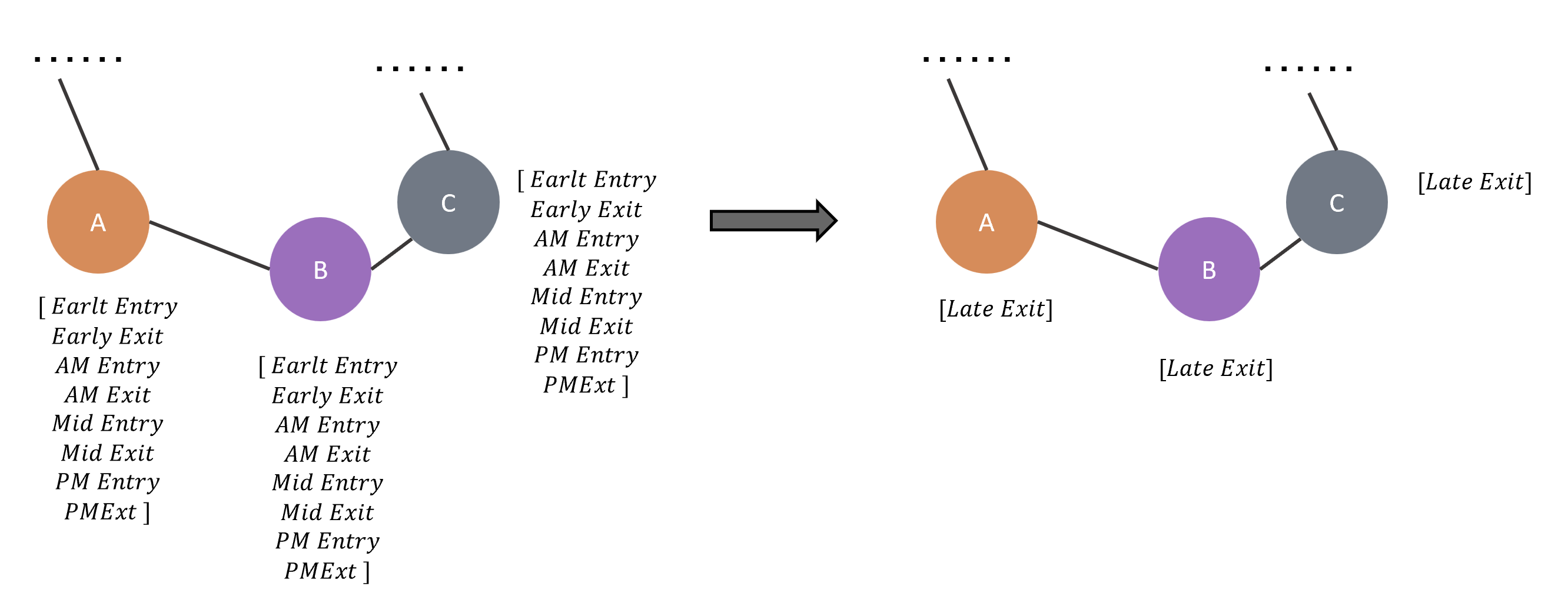}
  \caption{An example of late exit prediction. Station A, B and C predict late exit data based on all previous time stamps.
  }
  \label{fig:Prediction_example}
\end{figure}

The aim of this study is to predict the entry and exit data of each station at a specific time stamp, based on the entry and exit data from all other time stamps. To simplify the formulation process, the time stamps are indexed with numbers, which means \textbf{[Early, AM, Mid, PM, Late]} is regarded as $\mathbf{[1, 2, 3, 4, 5]}$. For example, the early entry data would be labeled as $X^1_{entry}$. The entry and exit data are predicted separately for optimal performance. To give an example, to predict the late entry data:
\[ [X^1_{entry},X^1_{exit},X^2_{entry},X^2_{exit},X^3_{entry},X^3_{exit},X^4_{entry},X^4_{exit}]\] 
\[\xRightarrow{\text{Predict}} [X^5_{entry}] \]
In total, to predict all the time stamps, 10 models will be trained for each year. \cref{fig:Prediction_example} gives an example of how to predict Late exit data.

\subsection{$K$-hop Graph}
The region the graph convolution operation could reach is called the receptive field. For metro flow prediction, the receptive field means the range of stations that can be used to gather information, which also means the length of journey that the passengers could possibly go through. In a real-life scenario, most of the passengers will travel for more than one station, which is contrary to the one-hop receptive field the inherent metro graph provides. Thus, it is necessary to increase the receptive field of this model. Building a deeper convolution network could increase the receptive field. However, most graph neural network structures have shallow structures \cite{quan2019brief}. When the receptive fields increases, over-smoothing could occur, which means the vertices within the same neighbourhood will have similar features \cite{li2018deeper}. Also, when the model becomes deeper, the gradient vanishing problem might occur, and the model becomes too complex, leading to possible overfitting.
To solve this problem, our study increases the receptive field by modifying the structure of the metro graph. Instead of the inherent Metro graph, the $K$-hop version of the graph will be used. Considering the adjacency matrix of the inherent metro graph is $A$, the $K$-hop is the graph constructed from $A^K$. An edge will connect two vertices if one can reach another within $K$ walks. Using the $K$-hop graph, the receptive field of a single layer increases from one to $k$. Thus we can build a shallower network and could avoid gradient vanishing and overfitting. The comparison of $k$-hop graph models and deep one-hop models will be delivered in the results section. As for the over-smoothing problem, the GraphSAGE model can help reduce the hazard. A detailed explanation of the over-smoothing problem and how GraphSAGE can solve it is in the next section. 

\begin{figure}[h]
     \centering
     \begin{subfigure}[htbp]{0.2\textwidth}
         \centering
         \includegraphics[height=1in]{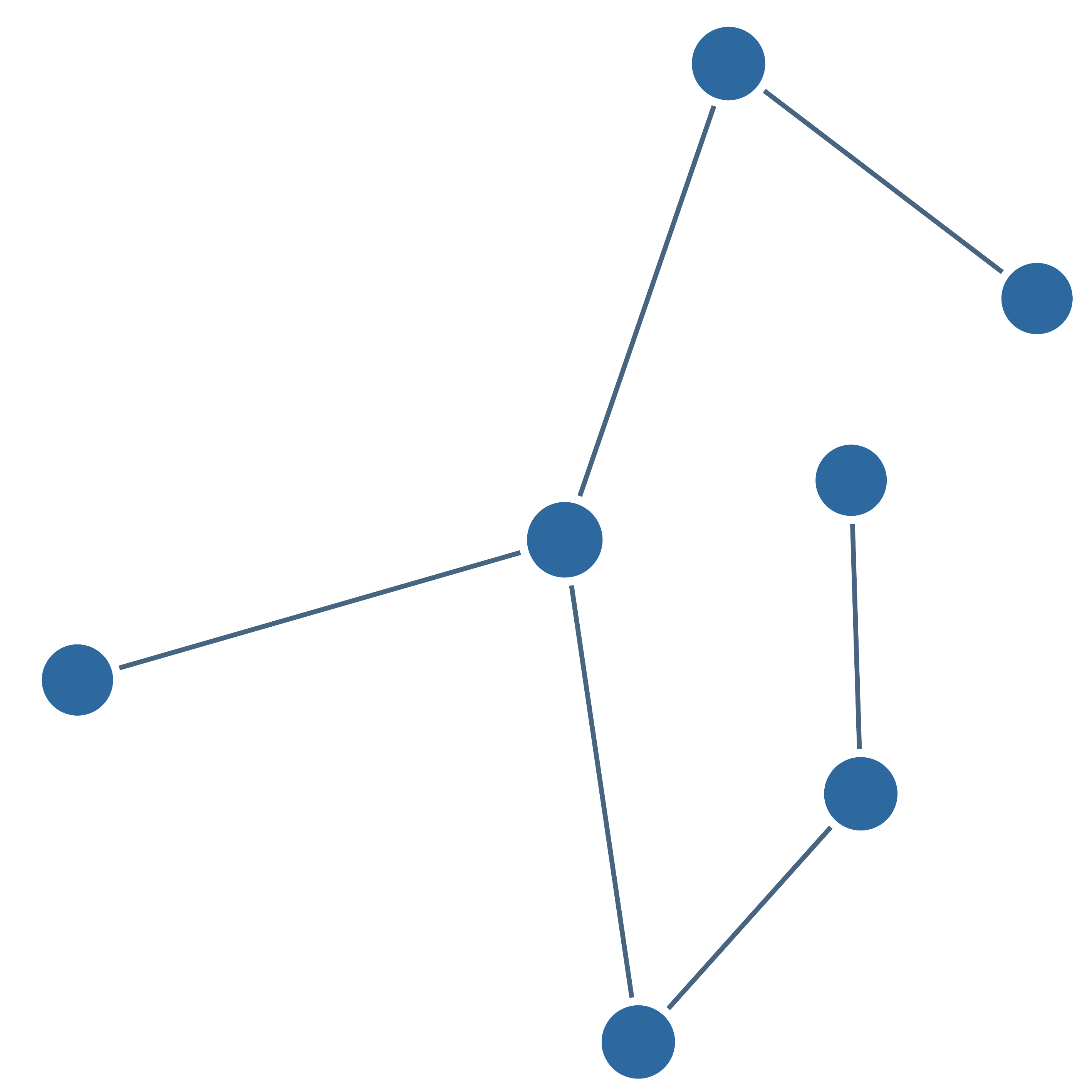}
         \subcaption{Original Graph}
     \end{subfigure}
     \begin{subfigure}[htbp]{0.2\textwidth}
         \centering
         \includegraphics[height=1in]{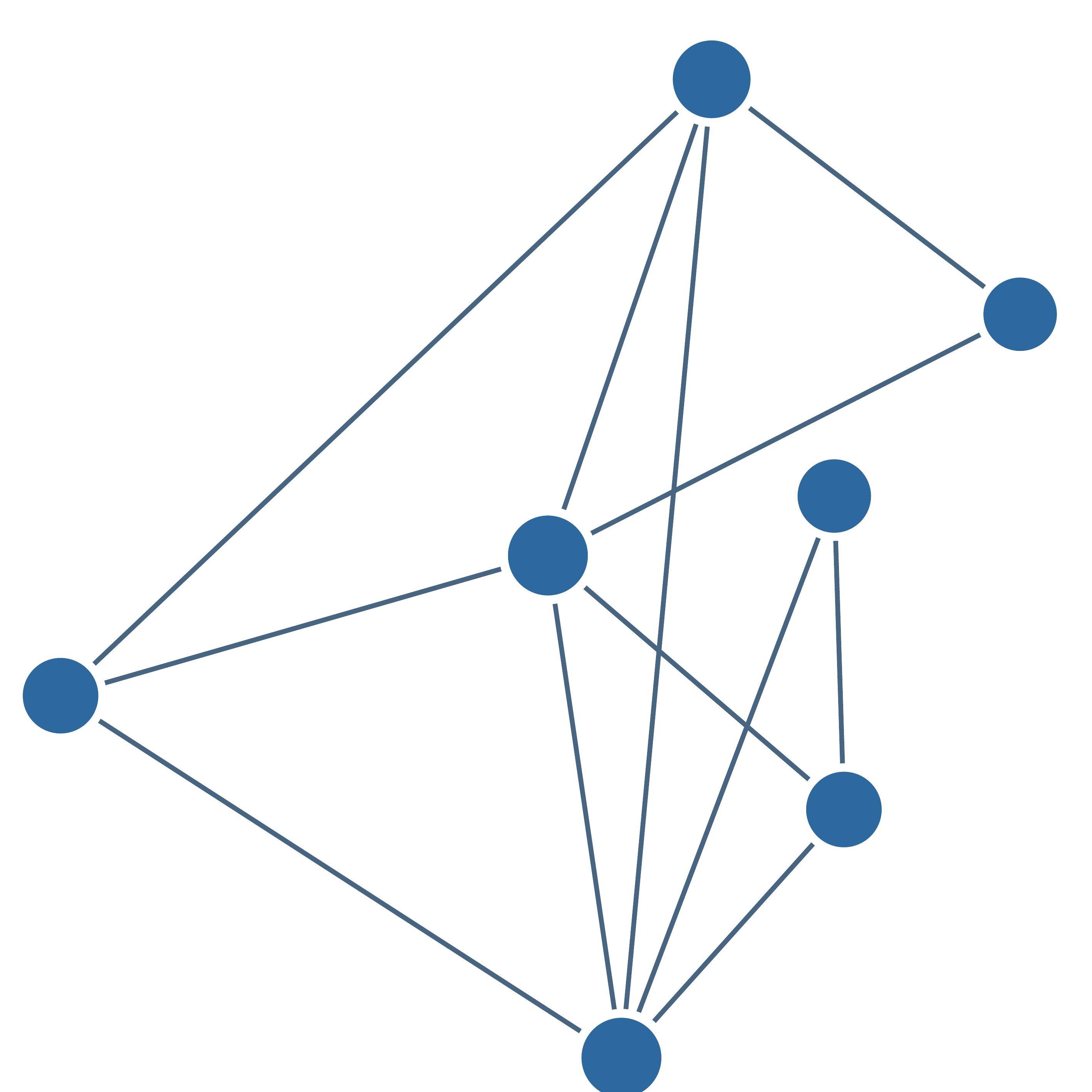}
         \subcaption{Two-hop Graph}
     \end{subfigure}
    \label{fig: k-hop example}
    \caption{An example of two-hop graph}
\end{figure}
 
\subsection{GraphSAGE}
\begin{figure}[h]
    \centering
    \includegraphics[width = 1 \linewidth]{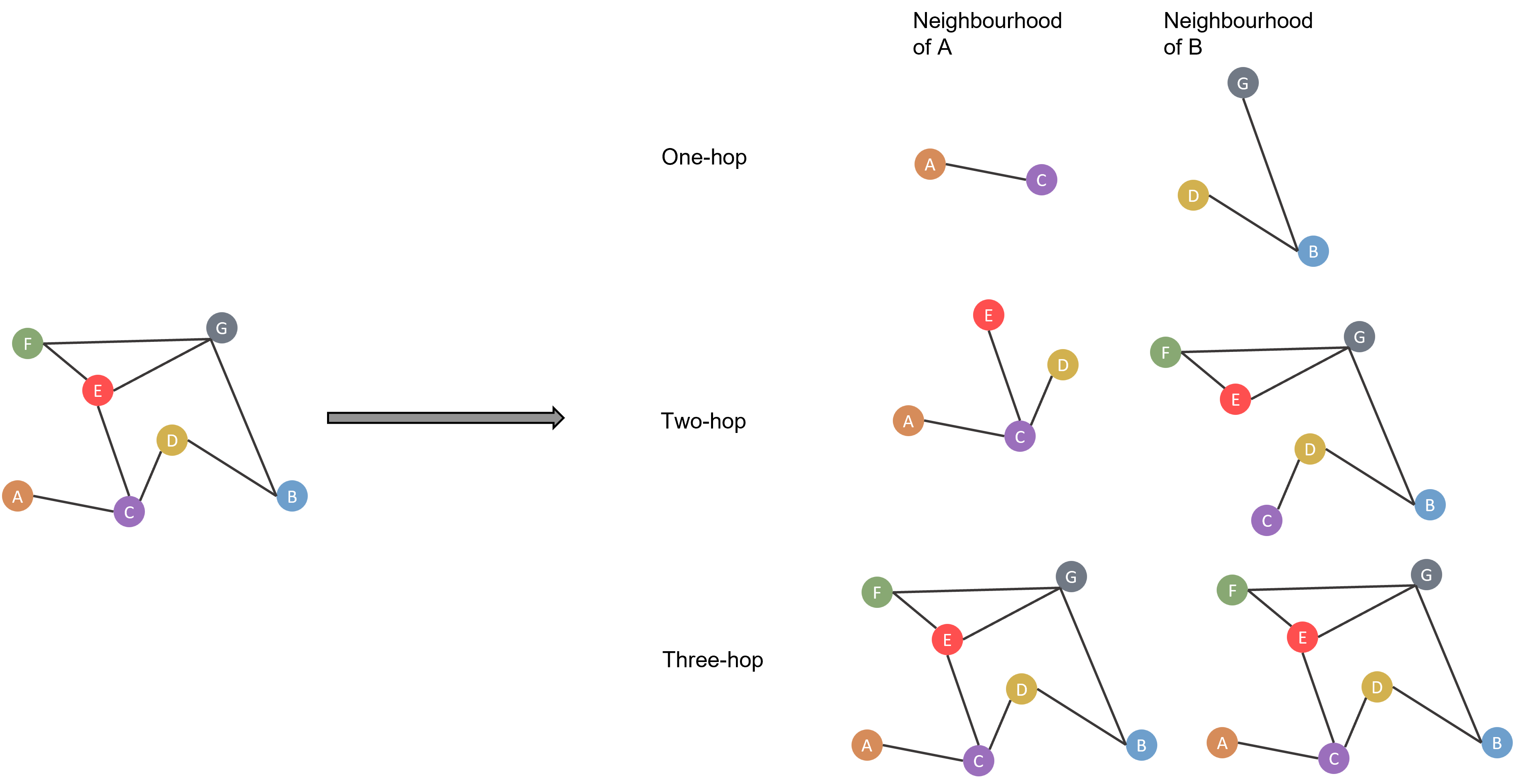}
    \caption{An illustration of the over-smoothing effect. It is manifest that the neighbourhoods of node A and B become more similar as the number of hops increases.}
    \label{fig:oversmoothing}
\end{figure}

For this study, GraphSAGE is used for convolution operation because GraphSAGE's sampling strategy reduces the over-smoothing problem caused by the $k$-hop graph \cite{hamilton2017inductive}.
Over-smoothing is a well-known issue for graph deep learning networks with a large receptive field. Graph deep learning algorithms usually contain two stages: aggregation and transformation. Each node aggregates information from its neighbours and transforms it into features as the new representation of itself. However, as the size receptive increases, the information aggregated from different nodes might overlap. Thus, different nodes will have similar representations and might result in a too-smooth graph with no distinction between nodes. \Cref{fig:oversmoothing} gives an example of over-smoothing. For a one-hop graph, node A and B has completely different neighbourhoods. When the hops increase to two, node A and B neighbourhoods become overlapped. Finally, node A and B neighbourhoods are identical for a three-hop graph.
The over-smoothing problem can be eased by introducing GraphSAGE. The concatenation procedure of GraphSAGE will project the feature of the node itself and the feature aggregated from neighbours onto separate feature spaces. Thus, the GraphSAGE algorithm can weight the self feature higher when over-smoothing starts to occur. An experiment is conducted to prove this idea, which will appear in the appendix. Below is a detailed introduction to the GraphSAGE algorithm.
Given a graph $G = (V, E)$, where $V$ stands for vertices and $E$ stands for edges. Now we assume there are $k$ aggregators ( denoted as $\text { AGGREGATE }_{k}, \forall k \in\{1, \ldots, K\}$ ), also a set of weights matrices $W^{k}, \forall k \in\{1, \ldots, K\}$, which are used as information transfer methods among layers. The following procedures shows the details of GraphSAGE learning process:
\begin{enumerate}
    \item Each node $v$ aggregates representation of nodes $h^k_{N(V)}$ of nodes from its one-hop neighbour $N(v)$, which depends on representation generated in the previous iteration. The initial representation is $h^0_{v} = x_{0}$, where $x_{0}$ is the input feature.
    \begin{equation}
    h_{N(v)}^{k}=\operatorname{AGGREGATE}_{k}\left(h_{u}^{k}, \forall u \in N(v)\right)
    \label{eq:Aggregation}
    \end{equation}
    \item The vertex's representation $h^{k-1}_{v}$ is concatenated with the aggregated neighbourhood vector $h^{k-1}_{N(v)}$. The concatenated representation is then passed through a fully-connected layer and a non-linear activation function $\sigma$ to generate the representation for next stage.
    \begin{equation}
    h_{v}^{k}=\sigma\left(W^{k} \cdot \operatorname{CONCAT}\left(h_{v}^{k-1}, h_{N(v)}^{k}\right)\right)
    \label{eq:CONCAT}
    \end{equation}
    \item The aggregator used in this study is the max-pooling aggregator. Representation of each neighbour of the vertex is fed to an individual fully-connected layer. Then a element wise max-pooling operator will aggregate the neighbour representation.
    
    \begin{multline}
    \text { AGGREGATE }_{k}^{\text {pool }} \\
    =\max \left(\left\{\sigma\left(W_{\text {pool }} \mathbf{h}_{u_{i}}^{k}+\mathbf{b}\right), \forall u_{i} \in \mathcal{N}(v)\right\}\right)
    \label{eq:Max pooling}
    \end{multline}
\end{enumerate}

\subsection{Hypergraph}
Till now all of our study is based on the traditional adjacency matrix denoted as $A \in \mathbb{R}^{N \times N}$, where $A_{ij} = 1$ if there is a connection between vertex $i$ and vertex $j$. Although the adjacency matrix could clearly indicate the physical connection between two stations, it lacks the ability to capture high-level connectivity information. Thus, it is necessary to introduce hypergraph to our study.

Hypergraph can be represented as a vertex set $V$ and a hyper edge set $E$. A hyper edge could contains any number of vertices. A hypergraph can be represented by an incidence matrix $M \in \mathbb{R}^{|V| \times |E|}$, where $M_{ve} = 1$ if $v \in e$. $|V|$ and $|E|$ stands for the number of vertices and hyper edges. A metro system can be considered as a hypergraph with each line as a hyper edge. Thus, the hypergraph can model the activity of changing between lines that the inherent metro graph cannot do.

However, the GraphSAGE model used in this study does not apply to hypergraph. To solve this problem, the hypergraph is expanded into graph using algorithm proposed by Zhou \cite{zhou2006learning}. The algorithm expands each edge into a clique. And the Laplacian is:

\begin{equation}
\Delta=\mathrm{D}_{V}^{-1 / 2} \mathrm{M}^{\top} \mathrm{ZD}_{E}^{-1} \mathrm{MD}_{V}^{-1 / 2}
\label{eq:clique expansion}
\end{equation}
where $M$ represents the incidence matrix. $ Z \in \mathbb{R}^{|E| \times |E|}$ is a diagonal matrix with weights of each hyper edge. ${D}_{V} \in \mathbb{R}^{|V| \times |V|}$ is a diagonal matrix with the node degrees and ${D}_{E} \in \mathbb{R}^{|E| \times |E|}$ is a diagonal matrix with the edge degrees.

The hypergraph extension could model the behaviour of line changing of travellers. But one major disadvantage coming with this expansion is the explosion of number of edges, since now stations within the same line is adjacent to each other. The large number of edges will cause over-smoothing thus undermine the performance. One way to improve this is to sample the edges. In this study, a percentage of  elements in the expanded adjacency matrix is randomly selected and set to zero to even out the huge edge numbers. It can be viewed as the Hadamard product of the expanded adjacency matrix $A$ and sample matrix $A_{sample}$

\begin{equation}
A_{sparsed}= A \odot A_{sample}
\label{eq:sampling}
\end{equation}

\begin{figure}[h]
    \centering
    \includegraphics[width = 1 \linewidth]{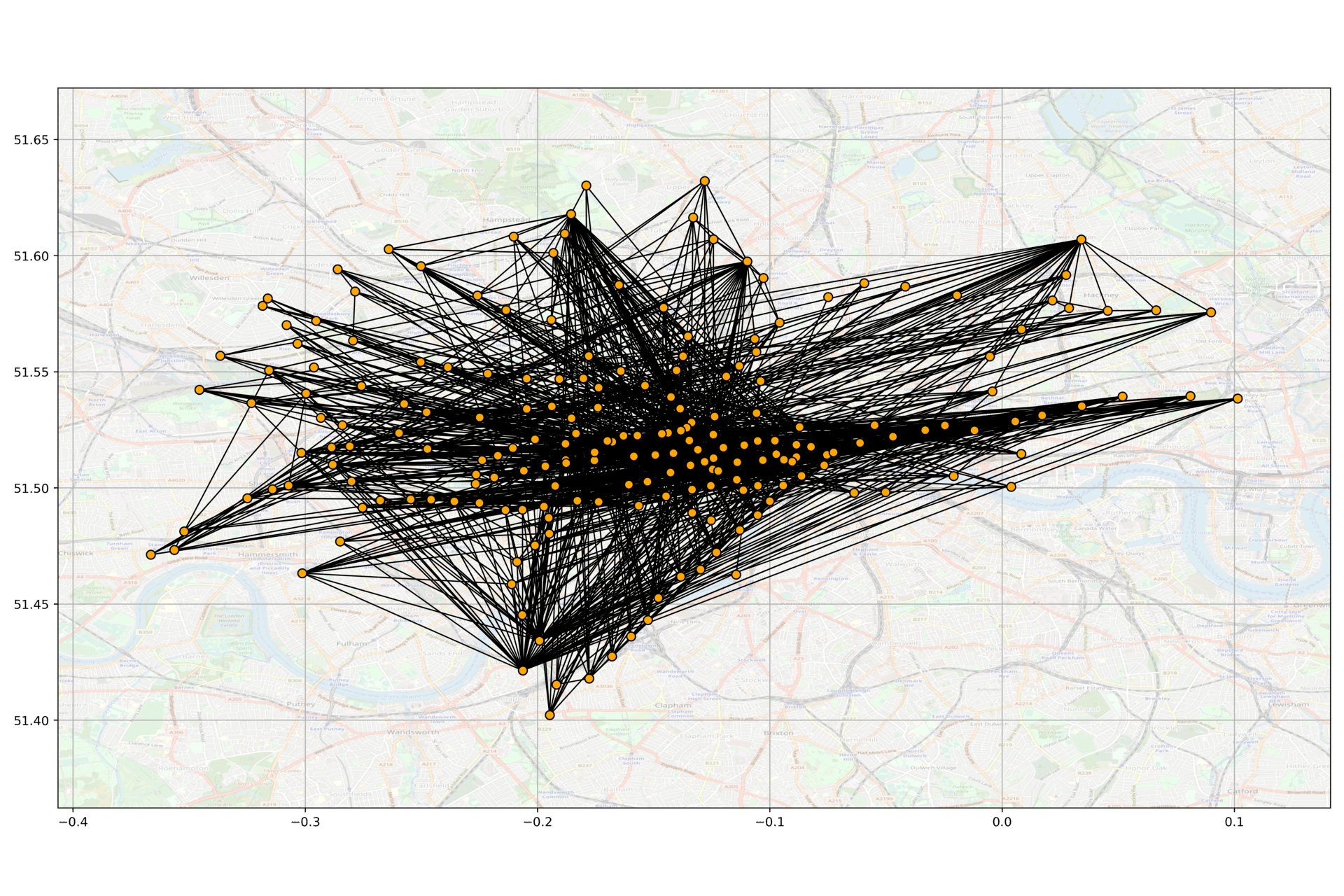}.
    \caption{The hypergraph after sampled with a sampling rate of 0.9}
    \label{fig:hypergraph}
\end{figure}

$A_{sample}$ has elements with values of zero and one. To utilise the hypergraph information, GraphSAGE modules mentioned before will be applied to the expanded and sampled hypergraph. The learned results can be concatenated with the output of the k-hop graph for information fusing \Cref{fig:hypergraph}. shows an example of the hypergraph with a sampling rate of 0.9.

\subsection{Real-world meaningful edge weights}
As reviewed in the above section, the edge weights scheme in traffic flow prediction can usually be divided into two categories:
\textbf{1} Directly use traffic-related information as edge weights (distance, ridership difference, etc.). \textbf{2} Learn from the dataset and traffic graph themselves (GAT, GATn2, etc.). Drawing inspiration from Bruno et al.'s work, which indicates that a station's in-flow and out-flow are related to its location's property (financial, business, residence, etc.). This study proposed a way of learning edge weights from social features.

Zone, average housing price and average expected life span of each station are used as social features. We proposed an MLP as an edge weights learner. The inputs of the edge weights learner are:
\begin{enumerate}
    \item The Zone difference between two stations
    \item The average housing price difference between two stations
    \item The average expected life span difference between two stations
\end{enumerate}
The learner can take the social features of any two stations and give the edge weights among them. Thus, the global problem of edge weights has now transferred to a local problem, which means only one MLP requires training. The trained MLP will work for any two stations. This is method could reduce the training times as well.

\subsection{Exploit temporal information}
\begin{figure}[h]
    \centering
    \includegraphics[width = 0.9 \linewidth]{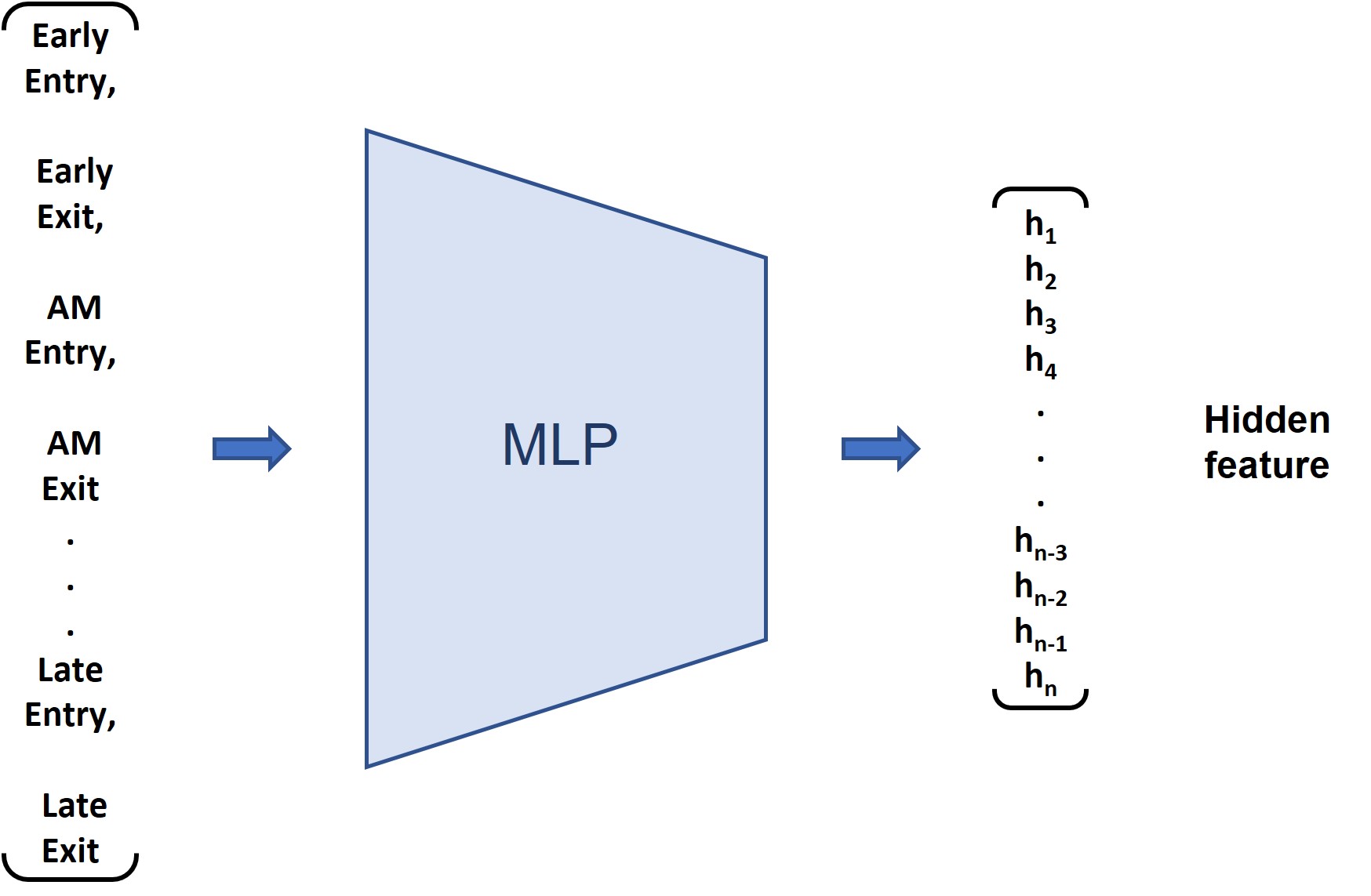}
    \caption{Exploitation of the temporal information.}
    \label{fig:temporal}
\end{figure}

All the proposed methods above aim to learn how to aggregate information from the neighbourhood to utilise spatial information. However, each station's feature also contains information in the temporal domain. A popular choice would be using models such as ARIMA, LSTM and Recurrent Neural Network to model the temporal dependencies. However, our data set lacks resolution in the time domain (only five time stamps), which makes it hard to apply these models to our dataset. For example, the rule of thumb for ARIMA models is to have at least 50, preferably more than 100 observations \cite{box1975intervention}. Thus, this study proposed an alternative way to utilise temporal information using an MLP. The temporal data at each station will be fed into an MLP to learn features. These learned features can be concatenated with the outputs of the k-hop graph, hypergraph or both to give a final prediction. \Cref{fig:temporal} gives an illustration of the temporal exploitation module.

\subsection{Overall architecture}
\begin{figure}[h]
  \includegraphics[width = 1 \linewidth]{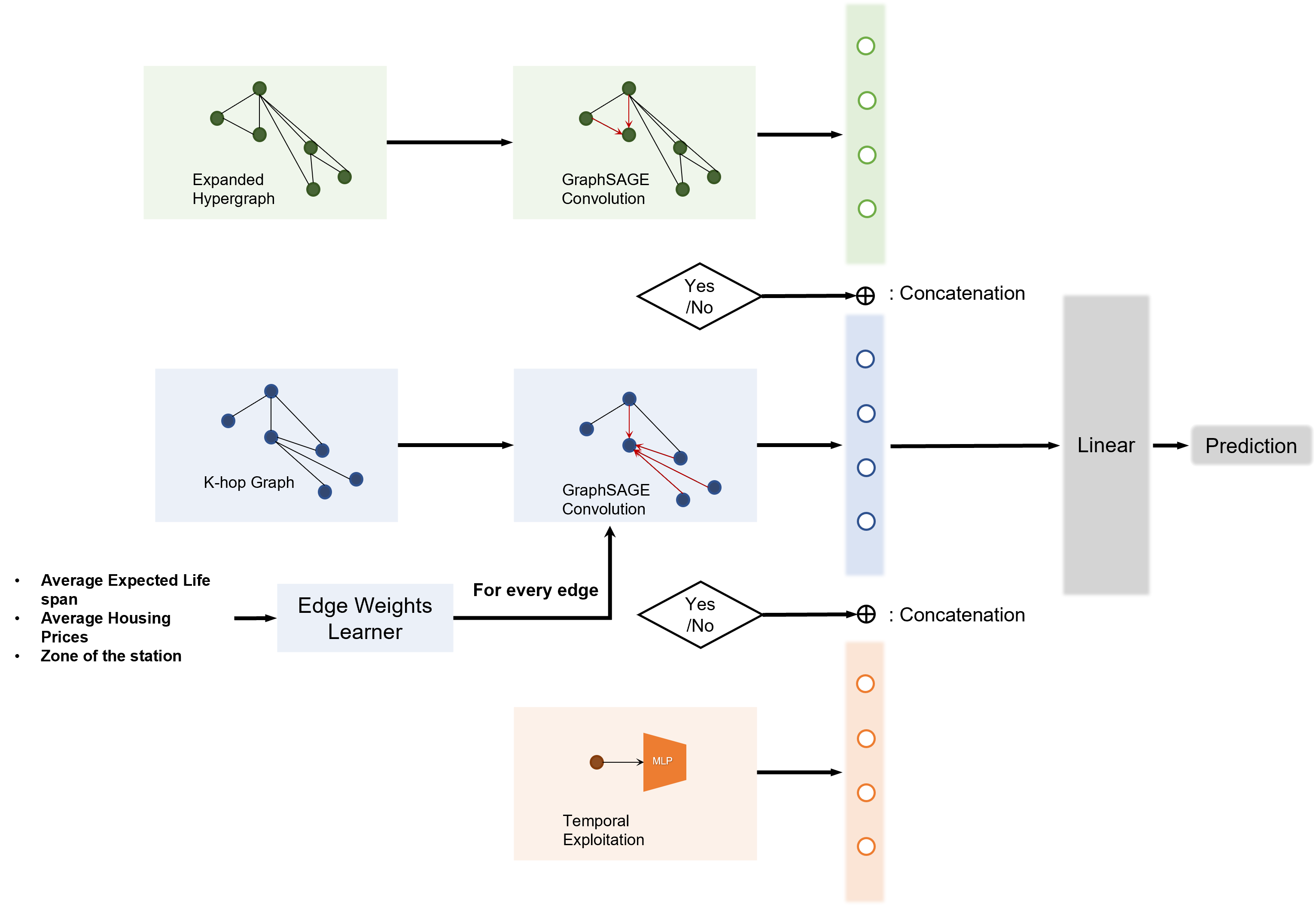}
  \caption{The overall structure of the proposed algorithm. The k-hop Graph is the main block in this study and the hypergraph and temporal exploitation modules are add-on  that can be connected or disconnected. 
  }
  \label{fig:architecture}
\end{figure}

\Cref{fig:architecture} shows the overall architecture of the proposed algorithm. The main body of the algorithm is the k-hop graph and the edge weights learner integrated with it. The temporal exploitation and the hypergraph modules can be regarded as add-ons. Once the module is added, the output of the added module will be concatenated with the output of the main body and go through a two-layer linear layer to give the final prediction. If no modules are added, the output of the main body itself will go through the linear layers to give the final prediction.

\section{Experiments}

\subsection{Experiment Design}
Thirteen years of data is randomly divided into a training set (nine years), a validation set (two years) and a test set (two years). The models are trained on the training set and tuned according to the performance of the validation set. The test set results are used to assess the performance of the models to give an unbiased result. Our study will use the GraphSAGE with $k$-hop graph and learned edge weights as the main structure, and then test the effect of adding or removing the temporal exploitation and hypergraph extensions. Also, the sampling rates of the hypergraph are 0.6, 0.7, 0.8 and 0.8. Any sampling rate below 0.6 has been tested and can only worsen the performance for all tasks. We will pin down some abbrevation for our models for easier understanding. 
\begin{itemize}
\item The combination of $k$-hop graph and edge weights learner will be denoted as \textbf{GraphSAGE + Edge weights learner}
\item The combination of $k$-hop graph, edge weights learner and temporal exploitation will be denoted as \textbf{KT}. 
\item The combination of $k$-hop, edge weights learner and hypergraph of sampling rate from 0.6 to 0.9 will be denoted as \textbf{KH 0.6},\textbf{KH 0.7},\textbf{KH 0.8} and \textbf{KH 0.9}. 
\item The combination of $k$-hop, edge weights learner, hypergraph and temporal exploitation will be denoted as: \textbf{KTH 0.6},\textbf{KTH 0.7},\textbf{KTH 0.8} and \textbf{KTH 0.9}.end
\end{itemize}
The proposed models will be trained to predict the entry and exit flow of all five time stamps (\textbf{Early, AM, Mid, PM, Late}). For each prediction task, the $k$ of the $k$-hop will vary from one to ten to find the hop that gives the best result. Traditional statistical methods such as ARIMA do not apply to our dataset since we only have a time sequence signal with a length of 5. Thus, they are not included in the comparison study. On the other hand, classic graph convolutional networks, including EdgeConv, TAG, GINConv, AGNNConv, Chebconv, GCN, GCN plus edge weights learner, GraphSAGE, GAT and GATv2Conv, are trained to give the benchmarking of this study. These models will be denoted as \textbf{comparison group}. So overall, 1900 models will be trained and compared (19 models $\times$ 5 time stamps $\times$ 2 direction of passenger flow $\times$ 10 hops ) 

\subsection{Model Setup}
\begin{figure*}
    \centering
    \includegraphics[width = 1 \textwidth]{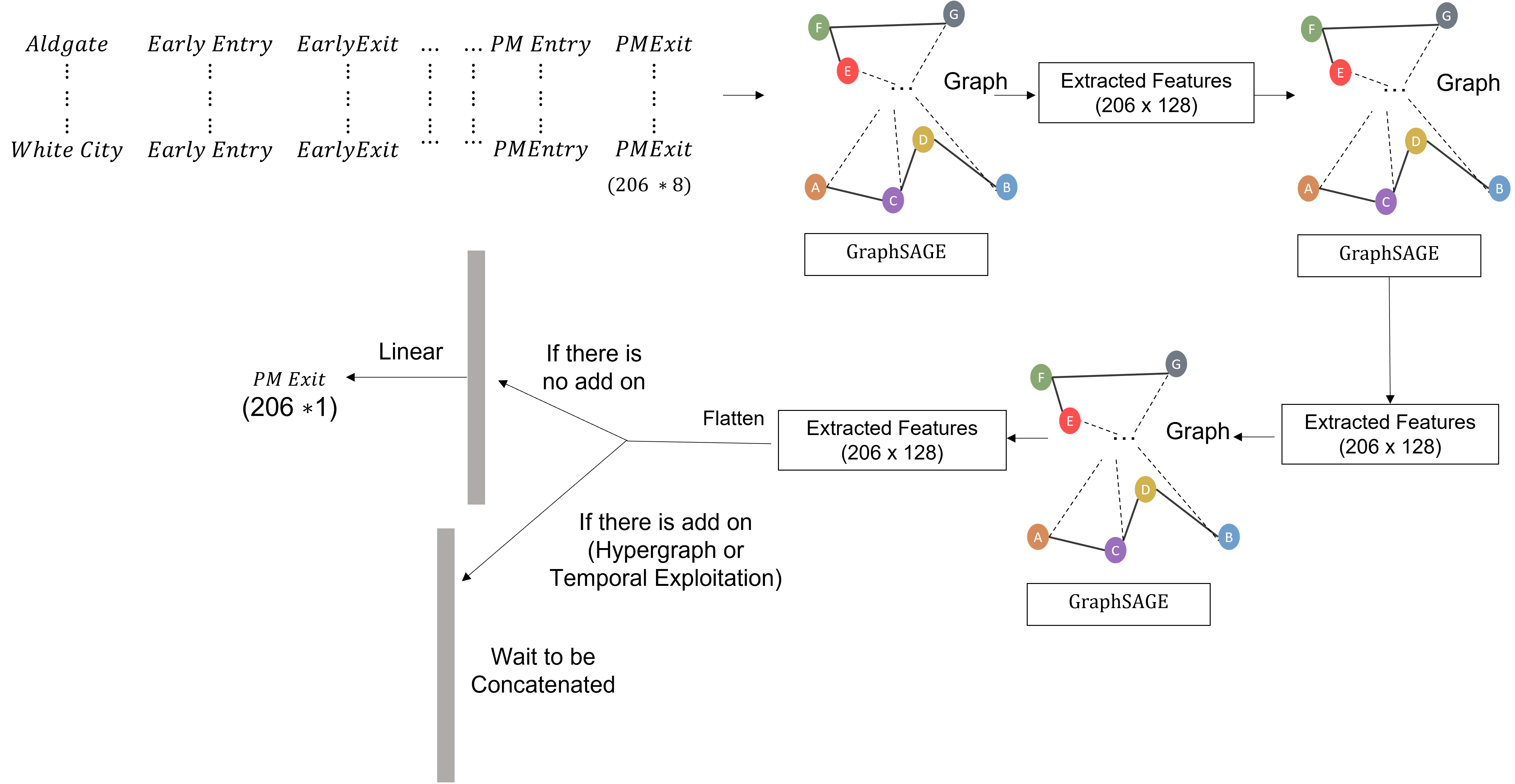}
    \caption{The GraphSAGE convolution module for proposed methods.}
    \label{fig:SAGE}
\end{figure*}

The GraphSAGE convolution layers in our proposed models has three layers. \cref{fig:SAGE} shows the structure of GraphSAGE module. AdamW is selected as the optimiser according to its practical performance. The learning rate starts from 0.005 and degrades with a ratio of 0.5 for every 200 epochs. Mean Absolute Percentage Error (MAPE) is used in this study as the evaluation metric since it clearly indicates how deviated the predicted results are. Moreover, the Mean Square Error (MSE) is used as the loss function metric since it gives the best results. The MAPE and MSE are defined as:

\begin{equation}
\begin{aligned}
\text { MAPE } &=\frac{1}{n} \sum_{i}^{n} \frac{\left|\widehat{Y}_{i}-Y_{i}\right|}{Y_{i}} \\
\text { MSE } &=\frac{1}{n} \sum_{i}^{n}\left|\widehat{Y}_{i}-Y_{i}\right|
\end{aligned}
\label{eq: Metrixs}
\end{equation}

where $\widehat{Y_{i}}$ is the predicted passenger flow and $Y_{i}$ is the ground truth passenger flow.

\subsection{Results and Discussions}
\begin{figure}[h]
    \centering
    \includegraphics[width = 1 \linewidth]{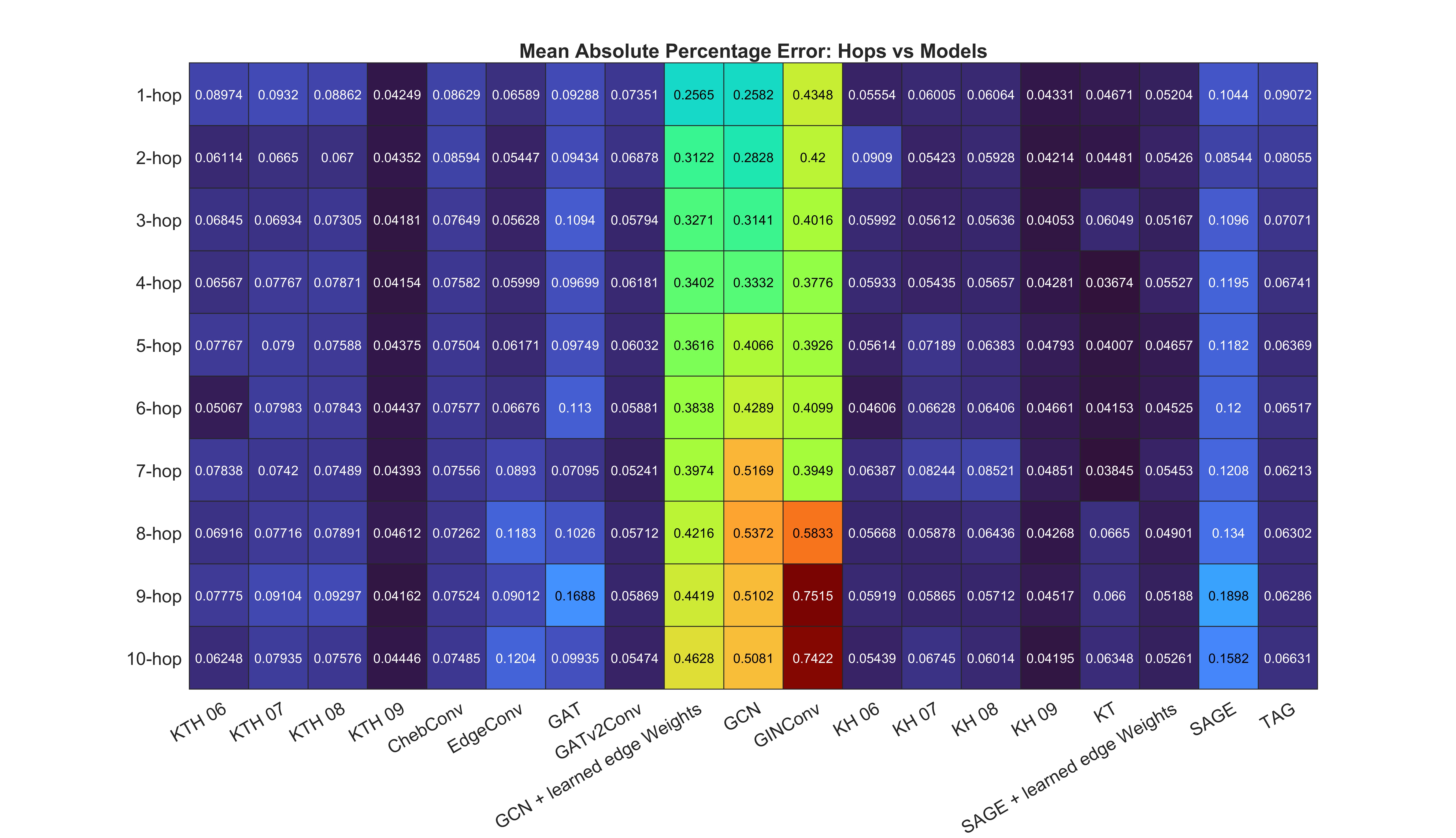}.
    \caption{The mean absolute percentage error of different hops and models for Mid Entry prediction}
    \label{fig:error surface}
\end{figure}

In this section, we will compare the results of different models and draw some conclusions. Figure \ref{fig:error surface} is an example and shows the mean absolute percentage error of the Mid Entry Prediction for different models from hop one to ten. \cref{tab:Early_entry,tab:Early_exit,tab:AM_entry,tab:AM_exit,tab:Mid_entry,tab:Mid_exit,tab:PM_entry,tab:PM_exit,tab:Late_entry,tab:Late_exit} present the best mean absolutely percentage errors with corresponding hop number for entry and exit prediction from early to late. We will analyse the results from a few perspectives. Firstly, we will regard the GraphSAGE plus edge weights learner and $k$-hop graph as the main body of our algorithm. We will break down the main body and see the influence that the $k$-hop graph and the edge weights learner can make. Then we will take the hypergraph and temporal exploitation modules as add-ons and discuss their performance. We will discuss the main body itself, the main body with hypergraph add-on, the main body with temporal exploitation add-on and the main body with both add-ons. Also, we will discuss if the GraphSAGE could reduce over-smoothing. Finally we will compare the $k$-hop scheme and the deep one-hop models to prove that our models works better even with the same receptive field.

\begin{itemize}
    \item The effect of $k$-hop 
    \item The effect of edge weights learner
    \item The result of the main body of the proposed algorithm: $k$-hop graph integrated with Edge weights learner
    \item The result with the temporal exploitation add-on
    \item The result with the Hypergraph add-on
    \item The result with both add-ons 
    \item The effect of GraphSAGE on over-smoothing
    \item $k$-hop versus deep one-hop networks
\end{itemize}

\subsubsection*{$k$-hop graph}
The $k$-hop graph is constructed to mimic the real-world passenger flow. $k$-hop graph grasp the social behaviour that metro passengers usually travel more than one stop. However, the $k$-hop graph also brings a problem: the explosion of edge numbers. This could be reflected from the best performances from \cref{tab:Early_entry,tab:Early_exit,tab:AM_entry,tab:AM_exit,tab:Mid_entry,tab:Mid_exit,tab:PM_entry,tab:PM_exit,tab:Late_entry,tab:Late_exit}. Some models, for example, GCN and EdgeConv, performs poorly when the number of hops increases in some time stamps. The problem for GCN and EdgeConv is that they cannot trim the number of edges and thus over-smoothing can happen.

\subsubsection*{The effect on over-smoothing using GraphSAGE}
The results also justify our previous point that GraphSAGE could reduce the inevitable over-smoothing coming with graph neural networks with large receptive fields. From \cref{tab:Early_entry,tab:Early_exit,tab:AM_entry,tab:AM_exit,tab:Mid_entry,tab:Mid_exit,tab:PM_entry,tab:PM_exit,tab:Late_entry,tab:Late_exit}, we can find our that the $k$ of average best performance $k$-hop graph for GCN and GraphSAGE is 1.5 and 2.7 respectively. Then considering that the model consists three convolution layers, the actually average best performance receptive field for GCN and GraphSAGE is 4.5 hops and 8.1 hops. Also GraphSAGE outperforms GCN in every tasks.  It is obvious that one can increase the receptive field without undermining the performance by using GraphSAGE .

\subsubsection*{Edge Weights Learner}
In this study, the GCN and GraphSAGE are used to compare the effect of edge weights learner. \cref{tab:SAGE_EL,tab:GCN_EL} shows the average improvement in MAPE of GraphSAGE and GCN model over hops after edge weights learner. A positive value means edge weight learner improves the performance. It is manifest that the edge weights learner increases the performance of models. For GCN, the edge weights learner increases the performance in most of the tasks. However, in two out of ten times stamps edge weights learner worsens its performance. However, there are significant and steady improvements for the GraphSAGE model. For all time stamps, the GraphSAGE plus edge weights learner method outperforms the GraphSAGE model alone by a significant amount. The results show that the edge weights learner could improve the performance generally.

\begin{table}[h!]
    \centering
    \begin{tabular}{c c c} 
    Time stamp   & MAPE \\ 
    Early Entry &8.21\% \\
    Early Exit &6.85\% \\   
    AM Entry &11.27\% \\
    AM Exit &13.11\% \\    
    Mid Entry &7.47\% \\
    Mid Exit &11.37\% \\
    PM Entry &10.99\% \\
    PM Exit &8.34\% \\
    Late Entry &6.57\% \\
    Late Exit &7.70\% \\
    \end{tabular}
    \caption{Average improvement in MAPE of GraphSAGE model over hops after edge weights learner.}
    \label{tab:SAGE_EL}
\end{table}

\begin{table}[h!]
    \centering
    \begin{tabular}{c c c} 
    Time stamp   & MAPE \\ 
    Early Entry &6.23\% \\
    Early Exit &3.59\% \\   
    AM Entry &1.23\% \\
    AM Exit &2.71\% \\    
    Mid Entry &3.91\% \\
    Mid Exit &-1.114\% \\
    PM Entry &3.41\% \\
    PM Exit &2.33\% \\
    Late Entry &-0.86\% \\
    Late Exit &0.69\% \\
    \end{tabular}
    \caption{Average improvement in MAPE of GCN model over hops after edge weights learner.}
    \label{tab:GCN_EL}
\end{table}
\subsubsection*{Main body ($k$-hop plus edge weights learner)}
Except for the AM Entry task, the main body outperforms all other models in the comparison group. It is manifest that GCN cannot handle the increasing hops and the peak of its performance occur at hop one or two. The hops corresponding to the main body's best performances for different tasks lie in the range from three to seven, which reveals the necessity of the $k$-hop idea. As for the edge weights scheme, the proposed main body outperforms GAT and GATv2Conv. The results indicate that based on social features, the introduced edge-weight scheme outperforms models that learn edge-weights from the data and graphs themselves. The performance increases significantly from GraphSAGE to the main body, which shows that the edge weights learner is better than uniform weight in prediction tasks.

\subsubsection*{Temporal Exploitation}
It is manifest that $\mathbf{KT}$ outperforms every model in all tasks except for PM Entry, including the comparison group and other proposed models. Owing to the small computation burden (one MLP for every station), $\mathbf{KT}$ performs well and steady. The performance improvement compared with the main body algorithm indicates that although each station only has five time stamps, there is still temporal information hidden in the time series data of each station.

\subsubsection*{Hypergraph}
Unlike the temporal exploitation extension, the hypergraph extension shows more instability. This study has tested four versions of hypergraph extension, with sampling percentage of 0.6, 0.7, 0.8 and 0.9. A weak pattern could be found that the performance increases with the sampling percentage. $\mathbf{KH 0.9}$ outperforms other sampling rates in 50\% of the tasks, and $\mathbf{KH 0.8}$ has the best results in 20\% of the tasks. Meanwhile, $\mathbf{KH 0.6}$ takes the rest 30\%.  $\mathbf{KH 0.7}$ is never the best performance model in any tasks. The added hypergraph extension could improve the performance of the main body in 60\% of the tasks. The $\mathbf{KH}$ shows instability mainly from a large number of edges even after sampling.

\subsubsection*{Temporal Exploitation + Hypergraph}
After adding the temporal exploitation extension, the $\mathbf{KTH}$ algorithm shows more stability. $\mathbf{KTH 0.9}$ outperforms $\mathbf{KTH}$ models with other sampling percentages in 70\% of the tasks meanwhile  $\mathbf{KTH 0.8}$ leads in 20\% of the tasks. The $\mathbf{KTH}$ outperforms the main body algorithm in 80\% of the tasks. By fusing the temporal information, the instability of the hypergraph extension is reduced.

\subsubsection*{$k$-hop graph model vs deep one-hop graph model}
As mentioned before, we stack $k$-hop graphs to form a shallow model instead of stack one-hop graphs to build a deep model for reaching a desired receptive field. Shallow network could help the gradient vanishing problem and reduce over-fitting. To justify our assumption, we pick the best performance model for each prediction task and convert them into deep one-hop models. \cref{tab:Structure} shows the difference in performance of shallow $k$-hop models and deep one-hop models. It is manifest that the $k$-hop model outperforms the one-hop models in all the tasks. 

\begin{table}[h!]
    \centering
    \begin{tabular}{c c c} 
    Task   & $k$-hop (MAPE) & one-hop (MAPE)\\ 
    Early Entry & 7.41\% & 16.21\% \\
    Early Exit & 10.66\% & 20.45\% \\
    AM Entry & 5.527\% & 8.03\% \\
    AM Exit & 5.8\% & 8.12\% \\
    Mid Entry & 3.67\% & 5.8\% \\
    Mid Exit & 4.66\% & 7.31\% \\
    PM Entry & 3.89\% & 4.22\% \\
    PM Exit & 3.86\% & 6.7\% \\
    Late Entry & 6.11\% & 9.46\% \\
    Late Exit & 5.73\% & 9.44\% \\
    
    \end{tabular}
    \caption{Different Structures}
    \label{tab:Structure}
\end{table}

\section{Conclusion and Discussion}
Focusing on improving the metro passenger flow prediction accuracy, this study proposed the \textbf{Hyper-GST} model. Our models can outperform the comparison group which consists of the most popular graph convolution algorithms. The proposed model could extract information from the $k$-hop graph, the hypergraph and the temporal domain. The model could mimic the real-world scenario where the passengers usually travel more than one stop, meanwhile also maintain a shallow structure to avoid possible gradient vanishing and overfitting. Also, the model could utilise the passengers' behaviour of changing metro line from the hypergraph module, which shows the potential of utilising hypergraph on the metro prediction task. The proposed model also learn edge weights not based on the common schemes like travel distance, but from socially meaningful parameters including the house price, expected life span and the traffic Zone the station is at. Moreover, from this study, we found that the GraphSAGE algorithm could prevent over-smoothing due to its separate projection approach.

\clearpage

\begin{table}[h!]
    \centering
    \begin{tabular}{c c c} 
    Model   & MAPE & Hop\\ 
    GINConv &54.89\% & 10\\
    GCN + Learned weights &33.77\% & 1\\
    GCN &28.79\% & 2\\
    ChebConv &18.96\% & 5\\
    TAG &15.57\% & 5\\
    GraphSAGE &13.40\% & 4\\
    GAT &12.56\% & 10\\
    EdgeConv &10.87\% & 3\\
    KTH 0.6 &9.38\% & 5\\
    KTH 0.7 &9.31\% & 8\\
    KTH 0.8 &9.26\% & 8\\
    KH 0.6 &9.22\% & 10\\
    KH 0.8 &9.12\% & 4\\
    KH 0.7 &9.00\% & 4\\
    GATv2Conv & 8.84\% & 6\\
    KH 0.9 &8.87\% & 10\\
    KTH 0.9 &8.83\% & 5\\
    GraphSAGE + Edge weights learner &7.46\% & 5\\
    KT &7.41\% & 7\\
    \end{tabular}
    \caption{Early Entry}
    \label{tab:Early_entry}
\end{table}

\begin{table}[h!]
    \centering
    \begin{tabular}{c c c} 
    Model   & MAPE & Hop\\ 
    GINConv &55.36\% & 9\\
    GCN + Learned weights &43.01\% & 4\\
    GCN &28.79\% & 2\\
    ChebConv &22.57\% & 4\\
    TAG &20.14\% & 5\\
    GAT &17.88\% & 8\\
    EdgeConv &13.70\% & 1\\
    GATv2Conv & 13.32\% & 5\\
    KH 0.6 &13.13\% & 2\\
    GraphSAGE &12.80\% & 4\\
    KH 0.8 &12.81\% & 1\\
    KH 0.7 &12.67\% & 7\\
    KTH 0.6 &12.42\% & 2\\
    KTH 0.8 &12.01\% & 2\\
    KTH 0.7 &11.59\% & 2\\
    KH 0.9 &11.18\% & 9\\
    KTH 0.9 &11.03\% & 2\\
    GraphSAGE + Edge weights learner &10.99\% & 6\\
    KT &10.66\% & 8\\
    \end{tabular}
    \caption{Early Exit}
    \label{tab:Early_exit}
\end{table}

\begin{table}[h!]
    \centering
    \begin{tabular}{c c c} 
    Model   & MAPE & Hop\\ 
    GINConv &49.59\% & 1\\
    GCN + Learned weights &27.77\% & 2\\
    GCN &23.65\% & 2\\
    GraphSAGE &10.67\% & 5\\
    GAT &10.30\% & 8\\
    ChebConv &9.53\% & 3\\
    TAG &7.66\% & 5\\
    GATv2Conv & 7.15\% & 5\\
    KH 0.9 &5.88\% & 2\\
    GraphSAGE + Edge weights learner &5.83\% & 4\\
    KH 0.7 &5.79\% & 9\\
    EdgeConv &5.68\% & 3\\
    KH 0.6 &5.68\% & 6\\
    KTH 0.7 &5.59\% & 5\\
    KTH 0.9 &5.59\% & 2\\
    KH 0.8 &5.53\% &74\\
    KTH 0.8 &5.46\% & 5\\
    KTH 0.6 &5.33\% & 4\\
    KT &5.527\% & 4\\
    \end{tabular}
    \caption{AM Entry}
    \label{tab:AM_entry}
\end{table}

\begin{table}[h!]
    \centering
    \begin{tabular}{c c c} 
    Model   & MAPE & Hop\\ 
    GINConv &50.53\% & 4\\
    GCN + Learned weights &35.54\% & 1\\
    GCN &23.65\% & 2\\
    GAT &10.44\% & 1\\
    ChebConv &10.43\% & 8\\
    GraphSAGE &10.24\% & 1\\
    TAG &9.44\% & 4\\
    KTH 0.8 &8.45\% & 6\\
    KTH 0.7 &8.36\% & 8\\
    KTH 0.6 &8.04\% & 2\\
    GATv2Conv & 7.82\% & 9\\
    KH 0.6 &7.24\% & 9\\
    EdgeConv &7.28\% & 1\\
    KH 0.8 &6.72\% & 4\\
    KH 0.7 &6.68\% & 4\\
    KH 0.9 &6.07\% & 10\\
    GraphSAGE + Edge weights learner &6.06\% & 7\\
    KTH 0.9 &5.94\% & 2\\ 
    KT &5.8\% & 4\\
    \end{tabular}
    \caption{AM Exit}
    \label{tab:AM_exit}
\end{table}

\begin{table}[h!]
    \centering
    \begin{tabular}{c c c} 
    Model   & MAPE & Hop\\ 
    GINConv &37.76\% & 4\\
    GCN &25.82\% & 1\\
    GCN + Learned weights &25.65\% & 1\\
    GraphSAGE &8.54\% & 2\\
    ChebConv &7.26\% & 8\\
    GAT &7.09\% & 7\\
    KTH 0.8 &6.70\% & 2\\
    KTH 0.7 &6.65\% & 2\\
    TAG &6.21\% & 7\\
    KH 0.8 &5.64\% & 3\\
    EdgeConv &5.45\% & 2\\
    KH 0.7 &5.42\% & 2\\
    GATv2Conv & 5.24\% & 7\\
    KTH 0.6 &5.07\% & 6\\
    KH 0.6 &4.61\% & 6\\
    GraphSAGE + Edge weights learner &4.53\% & 6\\
    KTH 0.9 &4.15\% & 4\\
    KH 0.9 &4.05\% & 3\\
    KT &3.67\% & 4\\
    
    \end{tabular}
    \caption{Mid Entry}
    \label{tab:Mid_entry}
\end{table}

\begin{table}[h!]
    \centering
    \begin{tabular}{c c c} 
    Model   & MAPE & Hop\\ 
    GINConv &41.05\% & 4\\
    GCN + Learned weights &27.44\% & 1\\
    GCN &25.82\% & 1\\
    GAT &8.77\% & 5\\
    GraphSAGE &8.47\% & 2\\
    ChebConv &8.00\% & 3\\
    TAG &7.63\% & 4\\
    KTH 0.7 &6.78\% & 1\\
    KTH 0.8 &6.47\% & 1\\
    KTH 0.6 &6.14\% & 8\\
    GATv2Conv & 5.80\% & 10\\
    KH 0.6 &5.67\% & 3\\
    EdgeConv &5.65\% & 2\\
    KH 0.7 &5.54\% & 4\\
    KH 0.8 &5.39\% & 4\\
    KH 0.9 &5.21\% & 8\\
    GraphSAGE + Edge weights learner &5.05\% & 3 \\
    KTH 0.9 &4.78\% & 3\\
    KT &4.66\% & 6\\
    \end{tabular}
    \caption{Mid Exit}
    \label{tab:Mid_exit}
\end{table}

\begin{table}[h!]
    \centering
    \begin{tabular}{c c c} 
    Model   & MAPE & Hop\\ 
    GINConv &46.55\% & 9\\
    GCN + Learned weights &29.11\% & 1\\
    GCN &26.06\% & 1\\
    GraphSAGE &10.56\% & 4\\
    GAT &8.76\% & 1\\
    ChebConv &8.02\% & 8\\
    TAG &6.69\% & 5\\
    EdgeConv &5.81\% & 5\\
    GATv2Conv & 5.40\% & 9\\
    GraphSAGE + Edge weights learner &4.77\% & 7 \\
    KTH 0.9 &4.46\% & 1\\
    KH 0.9 &4.35\% & 3\\
    KTH 0.6 &4.28\% & 8\\
    KTH 0.7 &4.27\% & 1\\
    KTH 0.8 &4.27\% & 1\\
    KT &4.24\% & 5\\
    KH 0.8 &4.04\% & 3\\
    KH 0.7 &4.01\% & 3\\
    KH 0.6 &3.89\% & 2\\
    \end{tabular}
    \caption{PM Entry}
    \label{tab:PM_entry}
\end{table}

\begin{table}[h!]
    \centering
    \begin{tabular}{c c c} 
    Model   & MAPE & Hop\\ 
    GINConv &39.88\% & 4\\
    GCN + Learned weights &26.85\% & 1\\
    GCN &26.06\% & 1\\
    GraphSAGE &10.28\% & 3\\
    GAT &8.07\% & 1\\
    ChebConv &7.40\% & 9\\
    KH 0.6 &6.89\% & 1\\
    KH 0.9 &6.75\% & 6\\
    TAG &6.32\% & 5\\
    EdgeConv &5.65\% & 5\\
    GATv2Conv & 5.21\% & 10\\
    GraphSAGE + Edge weights learner &5.09\% & 4 \\
    KH 0.7 &4.83\% & 6\\    
    KTH 0.7 &4.67\% & 9\\
    KTH 0.6 &4.62\% & 9\\
    KTH 0.8 &4.62\% & 9\\
    KH 0.8 &4.30\% & 6\\
    KTH 0.9 &4.26\% & 3\\
    KT &3.86\% & 6\\
    \end{tabular}
    \caption{PM Exit}
    \label{tab:PM_exit}
\end{table}

\begin{table}[h!]
    \centering
    \begin{tabular}{c c c} 
    Model   & MAPE & Hop\\ 
    GINConv &50.30\% & 4\\
    GCN + Learned weights &27.25\% & 1\\
    GCN &25.54\% & 1\\
    GAT &10.70\% & 7\\
    ChebConv &10.43\% & 8\\
    TAG &8.94\% & 8\\
    GraphSAGE &7.55\% & 1\\
    GATv2Conv & 7.45\% & 10\\
    EdgeConv &6.68\% & 3\\
    GraphSAGE + Edge weights learner &6.19\% & 5 \\
    KH 0.6 &7.07\% & 3\\
    KH 0.7 &6.83\% & 6\\
    KTH 0.6 &6.79\% & 4\\
    KH 0.8 &6.75\% & 6\\
    KTH 0.9 &6.62\% & 9\\
    KH 0.9 &6.34\% & 7\\
    KTH 0.7 &6.30\% & 4\\
    KTH 0.8 &6.30\% & 5\\
    KT &6.11\% & 7\\
    \end{tabular}
    \caption{Late Entry}
    \label{tab:Late_entry}
\end{table}

\begin{table}[h!]
    \centering
    \begin{tabular}{c c c} 
    Model   & MAPE & Hop\\ 
    GINConv &46.22\% & 4\\
    GCN + Learned weights &28.69\% & 1\\
    GCN &25.54\% & 2\\
    ChebConv &10.73\% & 10\\
    GAT &10.00\% & 1\\
    TAG &8.50\% & 8\\
    GraphSAGE &8.25\% & 1\\
    GATv2Conv & 6.98\% & 9\\
    EdgeConv &6.68\% & 3\\
    GraphSAGE + Edge weights learner &5.83\% & 10 \\
    KH 0.9 &10.56\% & 4\\
    KH 0.8 &10.43\% & 1\\
    KH 0.7 &10.35\% & 8\\
    KH 0.6 &10.03\% & 6\\
    KTH 0.6 &6.62\% & 3\\
    KTH 0.7 &5.97\% & 2\\
    KTH 0.8 &5.86\% & 7\\
    KTH 0.9 &5.75\% & 3\\
    KT &5.73\% & 6\\
    \end{tabular}
    \caption{Late Exit}
    \label{tab:Late_exit}
\end{table}

\FloatBarrier
\bibliographystyle{IEEEtran}
\bibliography{biblio.bib}

\begin{thebibliography}{10}
\providecommand{\url}[1]{#1}
\csname url@samestyle\endcsname
\providecommand{\newblock}{\relax}
\providecommand{\bibinfo}[2]{#2}
\providecommand{\BIBentrySTDinterwordspacing}{\spaceskip=0pt\relax}
\providecommand{\BIBentryALTinterwordstretchfactor}{4}
\providecommand{\BIBentryALTinterwordspacing}{\spaceskip=\fontdimen2\font plus
\BIBentryALTinterwordstretchfactor\fontdimen3\font minus
  \fontdimen4\font\relax}
\providecommand{\BIBforeignlanguage}[2]{{%
\expandafter\ifx\csname l@#1\endcsname\relax
\typeout{** WARNING: IEEEtran.bst: No hyphenation pattern has been}%
\typeout{** loaded for the language `#1'. Using the pattern for}%
\typeout{** the default language instead.}%
\else
\language=\csname l@#1\endcsname
\fi
#2}}
\providecommand{\BIBdecl}{\relax}
\BIBdecl

\bibitem{levin1980forecasting}
M.~Levin and Y.-D. Tsao, ``On forecasting freeway occupancies and volumes
  (abridgment),'' \emph{Transportation Research Record}, no. 773, 1980.

\bibitem{moayedi2008arima}
H.~Z. Moayedi and M.~Masnadi-Shirazi, ``Arima model for network traffic
  prediction and anomaly detection,'' in \emph{2008 international symposium on
  information technology}, vol.~4.\hskip 1em plus 0.5em minus 0.4em\relax IEEE,
  2008, pp. 1--6.

\bibitem{otoshi2015traffic}
T.~Otoshi, Y.~Ohsita, M.~Murata, Y.~Takahashi, K.~Ishibashi, and K.~Shiomoto,
  ``Traffic prediction for dynamic traffic engineering,'' \emph{Computer
  Networks}, vol.~85, pp. 36--50, 2015.

\bibitem{wang2018detection}
P.~Wang, L.~Li, Y.~Jin, and G.~Wang, ``Detection of unwanted traffic congestion
  based on existing surveillance system using in freeway via a cnn-architecture
  trafficnet,'' in \emph{2018 13th IEEE Conference on Industrial Electronics
  and Applications (ICIEA)}.\hskip 1em plus 0.5em minus 0.4em\relax IEEE, 2018,
  pp. 1134--1139.

\bibitem{yu2016data}
D.~Yu, Y.~Liu, and X.~Yu, ``A data grouping cnn algorithm for short-term
  traffic flow forecasting,'' in \emph{Asia-Pacific web conference}.\hskip 1em
  plus 0.5em minus 0.4em\relax Springer, 2016, pp. 92--103.

\bibitem{shao2021traffic}
Y.~Shao, Y.~Zhao, F.~Yu, H.~Zhu, and J.~Fang, ``The traffic flow prediction
  method using the incremental learning-based cnn-ltsm model: the solution of
  mobile application,'' \emph{Mobile Information Systems}, vol. 2021, 2021.

\bibitem{narmadha2021spatio}
S.~Narmadha and V.~Vijayakumar, ``Spatio-temporal vehicle traffic flow
  prediction using multivariate cnn and lstm model,'' \emph{Materials Today:
  Proceedings}, 2021.

\bibitem{ranjan2020city}
N.~Ranjan, S.~Bhandari, H.~P. Zhao, H.~Kim, and P.~Khan, ``City-wide traffic
  congestion prediction based on cnn, lstm and transpose cnn,'' \emph{IEEE
  Access}, vol.~8, pp. 81\,606--81\,620, 2020.

\bibitem{kipf2016semi}
T.~N. Kipf and M.~Welling, ``Semi-supervised classification with graph
  convolutional networks,'' \emph{arXiv preprint arXiv:1609.02907}, 2016.

\bibitem{hamilton2017inductive}
W.~Hamilton, Z.~Ying, and J.~Leskovec, ``Inductive representation learning on
  large graphs,'' \emph{Advances in neural information processing systems},
  vol.~30, 2017.

\bibitem{velivckovic2017graph}
P.~Veli{\v{c}}kovi{\'c}, G.~Cucurull, A.~Casanova, A.~Romero, P.~Lio, and
  Y.~Bengio, ``Graph attention networks,'' \emph{arXiv preprint
  arXiv:1710.10903}, 2017.

\bibitem{ali2022exploiting}
A.~Ali, Y.~Zhu, and M.~Zakarya, ``Exploiting dynamic spatio-temporal graph
  convolutional neural networks for citywide traffic flows prediction,''
  \emph{Neural networks}, vol. 145, pp. 233--247, 2022.

\bibitem{agafonov2020traffic}
A.~Agafonov, ``Traffic flow prediction using graph convolution neural
  networks,'' in \emph{2020 10th International Conference on Information
  Science and Technology (ICIST)}.\hskip 1em plus 0.5em minus 0.4em\relax IEEE,
  2020, pp. 91--95.

\bibitem{yang2019traffic}
B.~Yang, S.~Sun, J.~Li, X.~Lin, and Y.~Tian, ``Traffic flow prediction using
  lstm with feature enhancement,'' \emph{Neurocomputing}, vol. 332, pp.
  320--327, 2019.

\bibitem{cai2020noise}
L.~Cai, M.~Lei, S.~Zhang, Y.~Yu, T.~Zhou, and J.~Qin, ``A noise-immune lstm
  network for short-term traffic flow forecasting,'' \emph{Chaos: An
  Interdisciplinary Journal of Nonlinear Science}, vol.~30, no.~2, p. 023135,
  2020.

\bibitem{fang2021kalman}
W.~Fang, W.~Cai, B.~Fan, J.~Yan, and T.~Zhou, ``Kalman-lstm model for
  short-term traffic flow forecasting,'' in \emph{2021 IEEE 5th Advanced
  Information Technology, Electronic and Automation Control Conference
  (IAEAC)}, vol.~5.\hskip 1em plus 0.5em minus 0.4em\relax IEEE, 2021, pp.
  1604--1608.

\bibitem{mehdi2022entropy}
M.~Z. Mehdi, H.~M. Kammoun, N.~G. Benayed, D.~Sellami, and A.~D. Masmoudi,
  ``Entropy-based traffic flow labeling for cnn-based traffic congestion
  prediction from meta-parameters,'' \emph{IEEE Access}, vol.~10, pp.
  16\,123--16\,133, 2022.

\bibitem{nguyen2020eo}
T.~Nguyen, G.~Nguyen, and B.~M. Nguyen, ``Eo-cnn: an enhanced cnn model trained
  by equilibrium optimization for traffic transportation prediction,''
  \emph{Procedia Computer Science}, vol. 176, pp. 800--809, 2020.

\bibitem{li2021hybrid}
Y.~Li, S.~Chai, Z.~Ma, and G.~Wang, ``A hybrid deep learning framework for
  long-term traffic flow prediction,'' \emph{IEEE Access}, vol.~9, pp.
  11\,264--11\,271, 2021.

\bibitem{chen2018dynamic}
K.~Chen, F.~Chen, B.~Lai, Z.~Jin, Y.~Liu, K.~Li, L.~Wei, P.~Wang, Y.~Tang,
  J.~Huang \emph{et~al.}, ``Dynamic spatio-temporal graph-based cnns for
  traffic prediction,'' \emph{arXiv preprint arXiv:1812.02019}, 2018.

\bibitem{li2018graph}
J.~Li, H.~Peng, L.~Liu, G.~Xiong, B.~Du, H.~Ma, L.~Wang, and M.~Z.~A. Bhuiyan,
  ``Graph cnns for urban traffic passenger flows prediction,'' in \emph{2018
  IEEE SmartWorld, Ubiquitous Intelligence \& Computing, Advanced \& Trusted
  Computing, Scalable Computing \& Communications, Cloud \& Big Data Computing,
  Internet of People and Smart City Innovation
  (SmartWorld/SCALCOM/UIC/ATC/CBDCom/IOP/SCI)}.\hskip 1em plus 0.5em minus
  0.4em\relax IEEE, 2018, pp. 29--36.

\bibitem{huang2020lsgcn}
R.~Huang, C.~Huang, Y.~Liu, G.~Dai, and W.~Kong, ``Lsgcn: Long short-term
  traffic prediction with graph convolutional networks.'' in \emph{IJCAI},
  2020, pp. 2355--2361.

\bibitem{chen2020multitask}
Z.~Chen, B.~Zhao, Y.~Wang, Z.~Duan, and X.~Zhao, ``Multitask learning and
  gcn-based taxi demand prediction for a traffic road network,''
  \emph{Sensors}, vol.~20, no.~13, p. 3776, 2020.

\bibitem{chen2021graph}
S.~Chen, Y.~C. Eldar, and L.~Zhao, ``Graph unrolling networks: Interpretable
  neural networks for graph signal denoising,'' \emph{IEEE Transactions on
  Signal Processing}, vol.~69, pp. 3699--3713, 2021.

\bibitem{lu2020spatiotemporal}
B.~Lu, X.~Gan, H.~Jin, L.~Fu, and H.~Zhang, ``Spatiotemporal adaptive gated
  graph convolution network for urban traffic flow forecasting,'' in
  \emph{Proceedings of the 29th ACM International Conference on Information \&
  Knowledge Management}, 2020, pp. 1025--1034.

\bibitem{yu2020forecasting}
B.~Yu, Y.~Lee, and K.~Sohn, ``Forecasting road traffic speeds by considering
  area-wide spatio-temporal dependencies based on a graph convolutional neural
  network (gcn),'' \emph{Transportation research part C: emerging
  technologies}, vol. 114, pp. 189--204, 2020.

\bibitem{chen2020multi}
W.~Chen, L.~Chen, Y.~Xie, W.~Cao, Y.~Gao, and X.~Feng, ``Multi-range attentive
  bicomponent graph convolutional network for traffic forecasting,'' in
  \emph{Proceedings of the AAAI conference on artificial intelligence},
  vol.~34, no.~04, 2020, pp. 3529--3536.

\bibitem{data}
``vis,'' \url{https://github.com/oobrien/vis}, accessed: 2022-02-08.

\bibitem{quan2019brief}
P.~Quan, Y.~Shi, M.~Lei, J.~Leng, T.~Zhang, and L.~Niu, ``A brief review of
  receptive fields in graph convolutional networks,'' in \emph{IEEE/WIC/ACM
  International Conference on Web Intelligence-Companion Volume}, 2019, pp.
  106--110.

\bibitem{li2018deeper}
Q.~Li, Z.~Han, and X.-M. Wu, ``Deeper insights into graph convolutional
  networks for semi-supervised learning,'' in \emph{Thirty-Second AAAI
  conference on artificial intelligence}, 2018.

\bibitem{zhou2006learning}
D.~Zhou, J.~Huang, and B.~Sch{\"o}lkopf, ``Learning with hypergraphs:
  Clustering, classification, and embedding,'' \emph{Advances in neural
  information processing systems}, vol.~19, 2006.

\bibitem{box1975intervention}
G.~E. Box and G.~C. Tiao, ``Intervention analysis with applications to economic
  and environmental problems,'' \emph{Journal of the American Statistical
  association}, vol.~70, no. 349, pp. 70--79, 1975.

\end{thebibliography}


\clearpage
\section{Appendix}
\subsection*{GraphSAGE on Over-smoothing}
For the GraphSAGE algorithm, the feature of the node itself and the feature aggregated from its neighbours are concatenated and then transformed. The aggregation, concatenate and transform procedure can be rewritten as \cref{eq:CONCAT2}. The self-feature ($W^{k}_{1} \cdot h_{v}^{k-1}$) and neighbour-feature ($W^{k}_{2} \cdot h_{N(v)}^{k}$) can be considered to be on different feature space. Thus, when the over-smoothing occurs, $h_{N(v)}^{k}$ is similar for all nodes, the GraphSAGE algorithm can adjust the weights of $W^{k}_{1}$ and $W^{k}_{2}$ to lift the influence of self-feature. To prove this idea, an experiment is conducted.

\begin{equation}
    \begin{aligned}
	 & h_{N(v)}^{k}=\operatorname{AGGREGATE}_{k}\left(h_{u}^{k}, \forall u \in N(v)\right)\\
     &h_{v}^{k}= W^{k}_{1} \cdot h_{v}^{k-1} + W^{k}_{2} \cdot h_{N(v)}^{k}
    \end{aligned}
\label{eq:CONCAT2}
\end{equation}

The dataset has been reduced to 57 stations containing only Zone 1 stations. A 10-hops graph is generated so that 52 stations are adjacent to all other 56 stations. Two models are generated, which contain a graph convolution layer and an output layer. The first model has a GCN layer, and the second one has a GraphSAGE layer. To further address the over-smoothing issue, the mean aggregator is used for GraphSAGE. The mean aggregator will average all neighbour features and resemble the GCN aggregation procedure. The models are trained on 2003 data to predict PM entry data.

\begin{figure}
    \centering
    \includegraphics[width = 1 \linewidth]{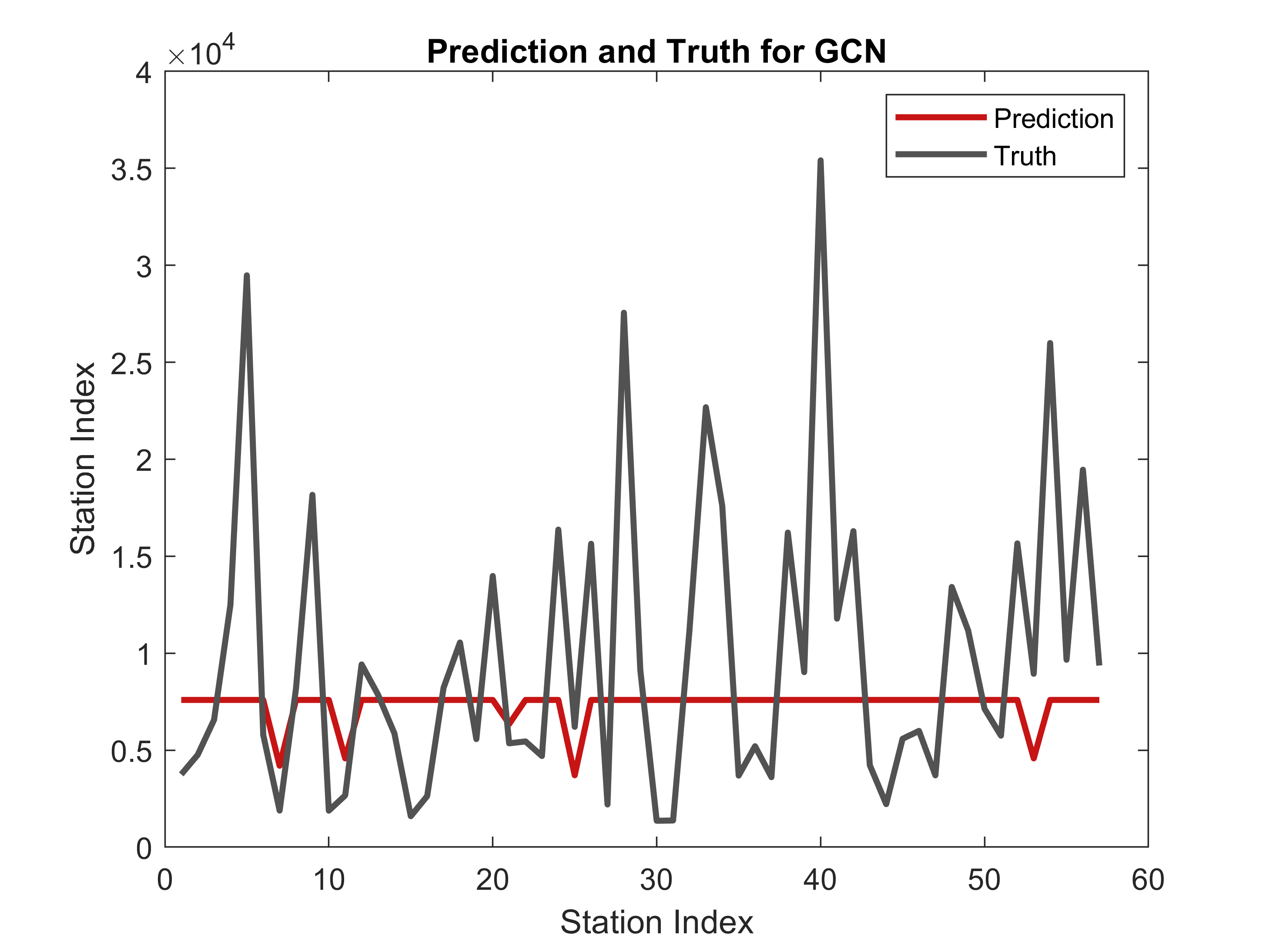}
    \caption{The prediction of GCN model and the ground truth.}
    \label{fig:GCN_result}
\end{figure}

\cref{fig:GCN_result} shows the prediction of the GCN model and the ground truth. It is manifest that the prediction is exactly the same for all nodes except for the five stations that are not connected to all other stations. \cref{eq:GCN_aggregate} illustrates the aggregation and transformation process for GCN. It is worth mentioning that the self-feature of the node is preserved by adding a self-loop as a common trick to avoid over-smoothing. Unlike GraphSAGE, the aggregation step fuses the feature of the node itself and the neighbour features. For this case, the 52 fully connected nodes will have the same representation, and all 57 nodes will have very similar features, which forces the model to give similar outputs and fail the prediction tasks.  

\begin{equation}
h_v^{(k)}=\sigma\left(b^{(k-1)}+\sum_{j \in \mathcal{N}(v)\cup v} \frac{1}{c_{j v}} h_j^{(k-1)} W^{(k-1)}\right)
\label{eq:GCN_aggregate}
\end{equation}

\begin{figure}
    \centering
    \includegraphics[width = 1 \linewidth]{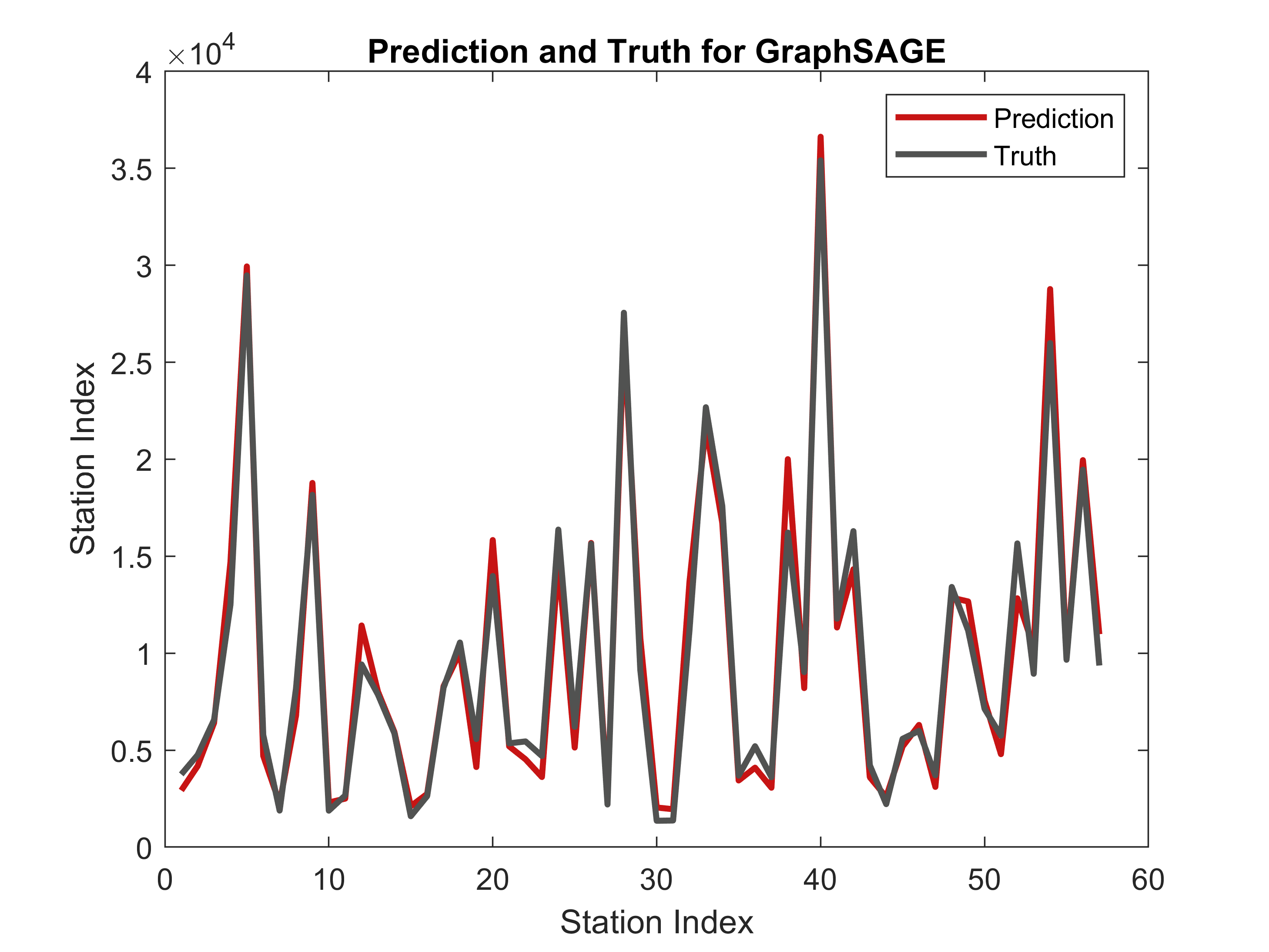}
    \caption{The prediction of GraphSAGE model and the ground truth.}
    \label{fig:SAGEresult}
\end{figure}

\begin{figure}
    \centering
    \begin{subfigure}[b]{1\linewidth}
    \includegraphics[width = 1\linewidth]{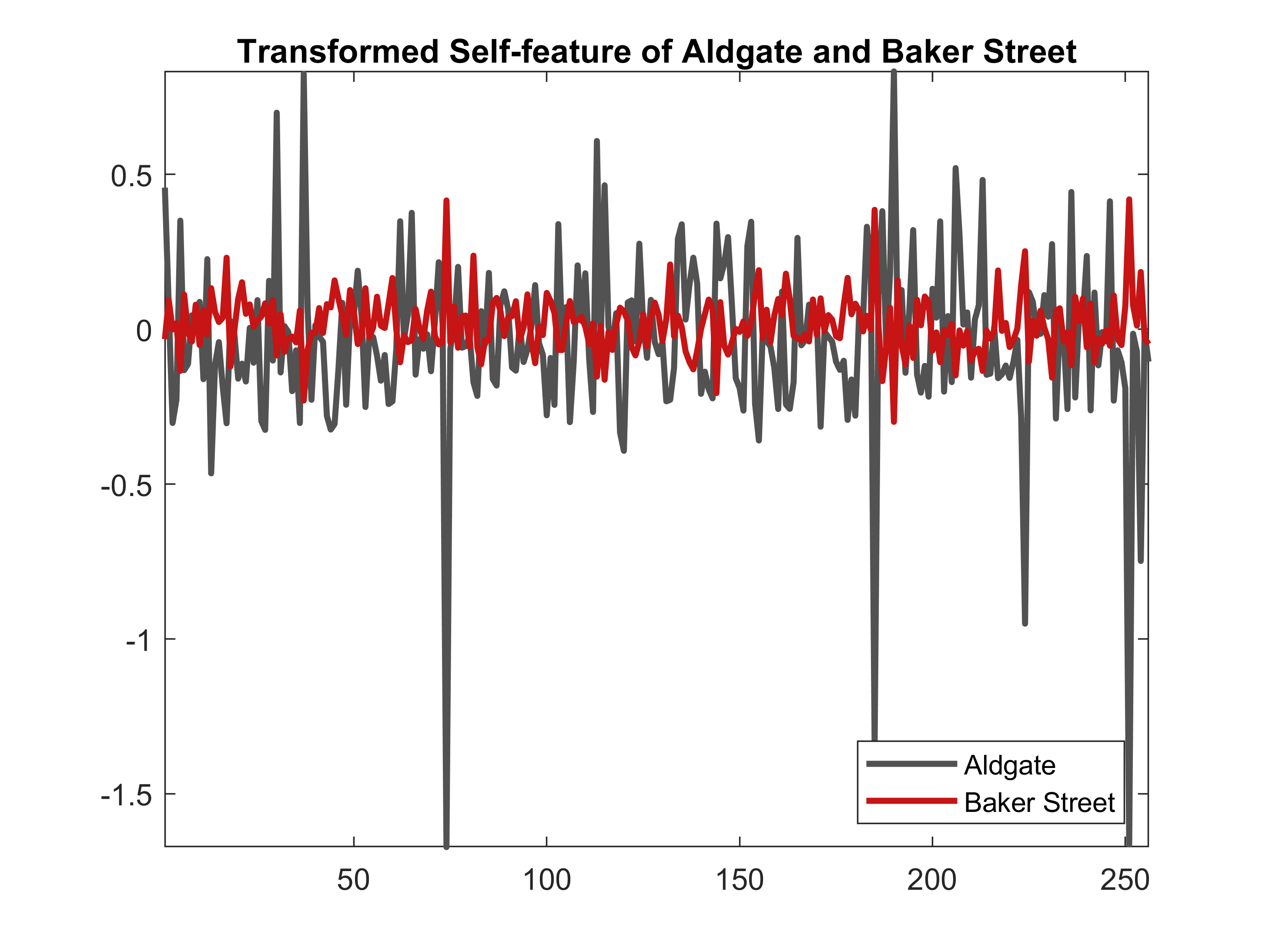}
    \caption{The Self-feature of Aldgate and Baker street plotted together.}
    \end{subfigure}
    \hfill
    \begin{subfigure}[b]{1\linewidth}
    \includegraphics[width = 1\linewidth]{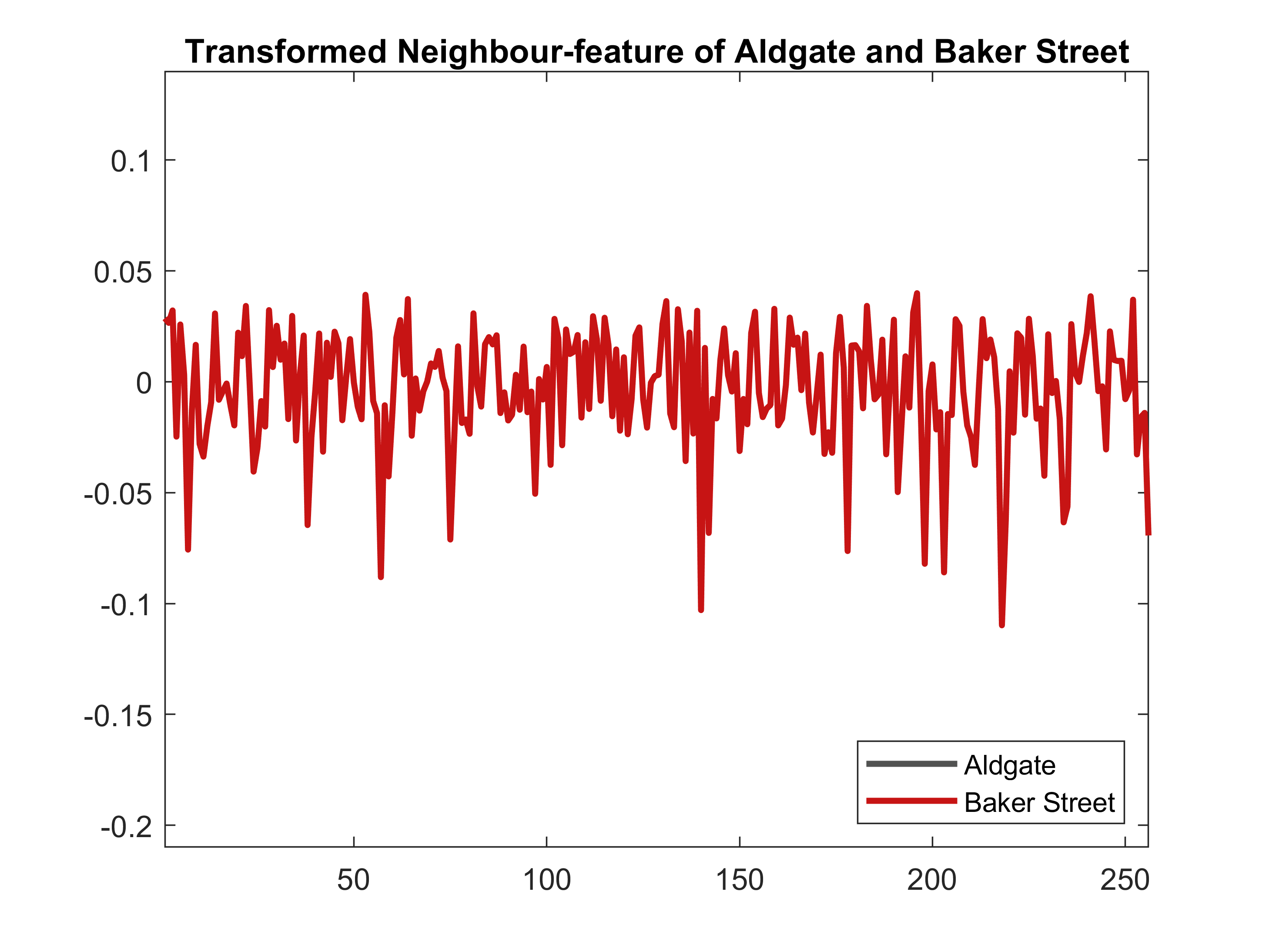}
    \caption{The Neighbour-feature of Aldgate and Baker street plotted together.}
    \end{subfigure}
    \caption{The different self-feature and same neighbour-feature of Aldgate and Baker Street.}
    \label{fig:Self/Neighbour feature}
\end{figure}

\begin{figure}
    \centering
    \begin{subfigure}[b]{1\linewidth}
    \includegraphics[width = 1\linewidth]{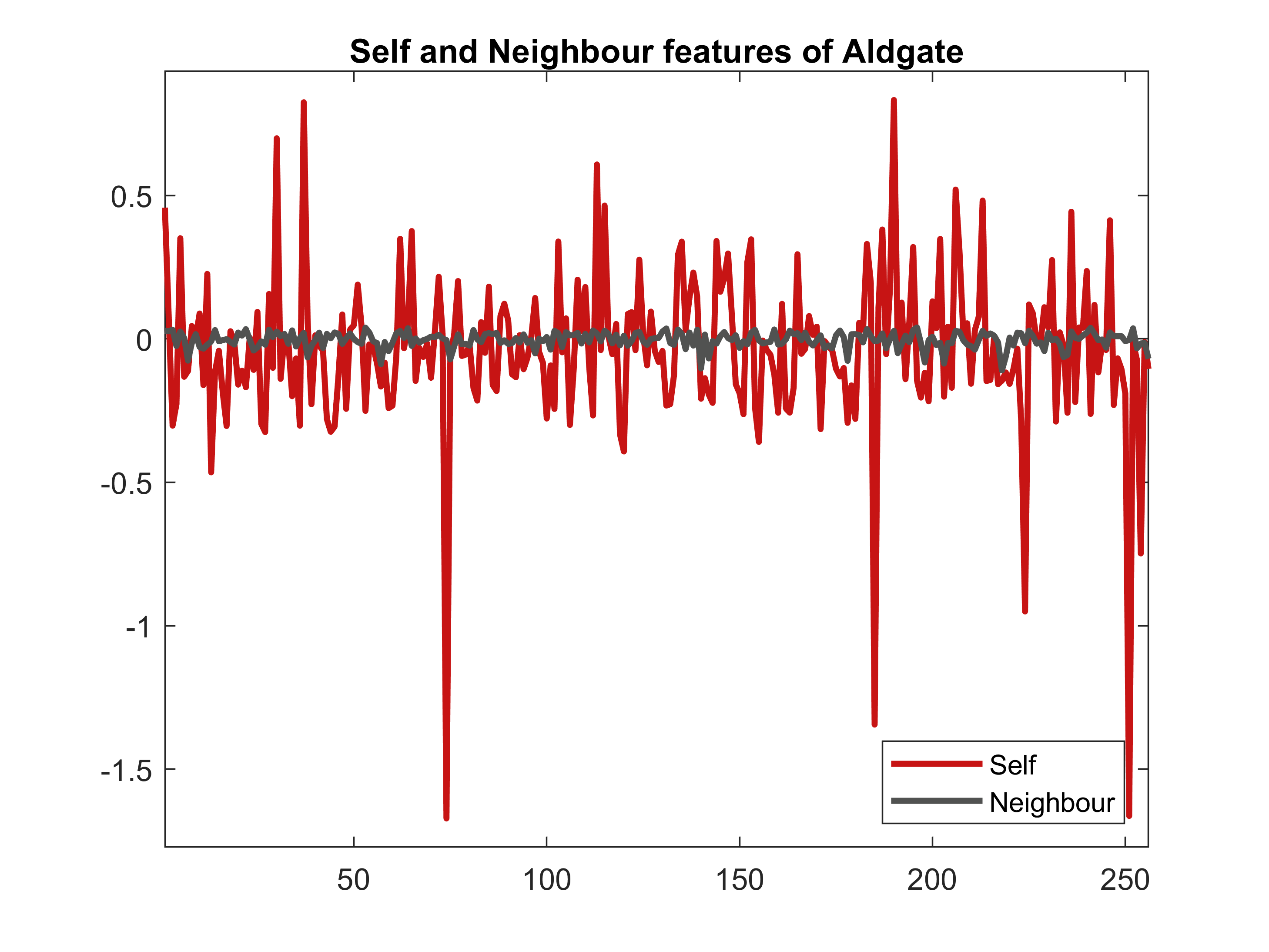}
    \caption{The Self-feature and neighbour-feature of Aldgate plotted together.}
    \end{subfigure}
    \hfill
    \begin{subfigure}[b]{1\linewidth}
    \includegraphics[width = 1\linewidth]{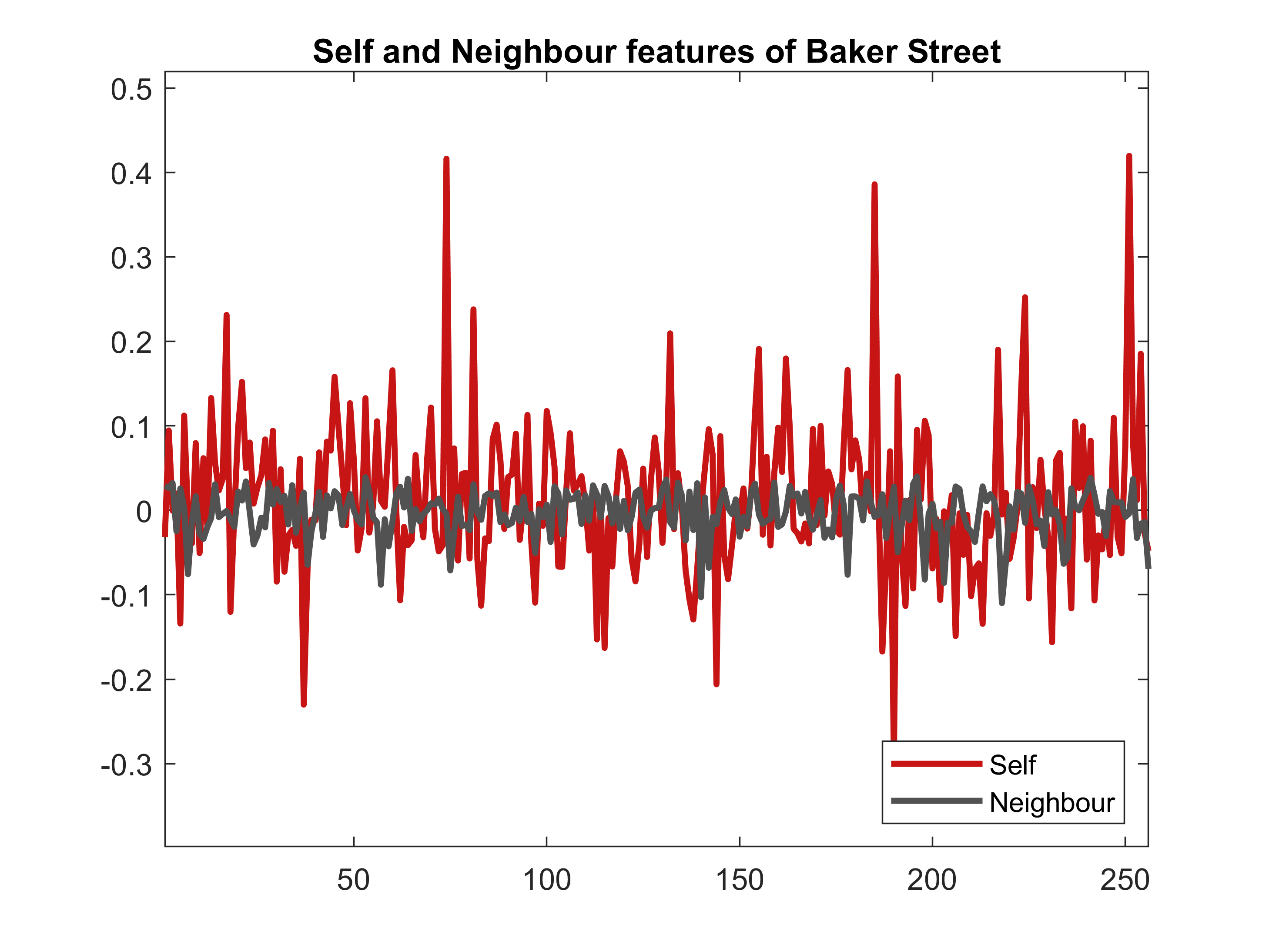}
    \caption{The Neighbour-feature of Aldgate and Baker street plotted together.}
    \end{subfigure}
    \caption{A more direct representation of how GraphSAGE weight down the over-smoothing neighbour-features}
    \label{fig:feature_together}
\end{figure}

\cref{fig:SAGEresult} shows the prediction of GraphSAGE and the ground truth. The performance of GraphSAGE is obviously better than the performance of the GCN model. The outputs of the GraphSAGE convolution layer of 'Aldgate' and 'Baker Street stations are investigated to investigate this performance increase further. \cref{fig:Self/Neighbour feature} shows the neighbour-feature and self-feature of Aldgate and Baker street, respectively. It is clear that the self-features, $W^{k}_{1} \cdot h_{v}^{k-1}$, is different for Aldgate and Baker street. Meanwhile, due to over-smoothing, the neighbour-features, $ W^{k}_{2} \cdot h_{N(v)}^{k}$, are exactly the same for both stations. What is more, the absolute magnitude of the self-feature is larger than that of the neighbour-feature, which can be observed as more directive through \cref{fig:feature_together}. By projecting the self and neighbour feature into different spaces, the GraphSAGE could automatically weigh down the neighbour-features which causes the over-smoothing, thus maintaining each node's distinct features and reducing over-smoothing.

\end{document}